\documentclass[lettersize,journal]{IEEEtran}
\usepackage{amsmath,amsfonts}
\usepackage{array}
\usepackage{textcomp}
\usepackage{stfloats}
\usepackage{url}
\usepackage{verbatim}
\usepackage{graphicx}
\usepackage{cite}

\usepackage{xcolor}

\usepackage{pgfplots} 
\usepackage{subcaption}
\usepackage{algorithm}
\usepackage{algpseudocode}
 \usepackage{multirow}
 \usepackage{threeparttable}
 \usepackage{makecell}

\usepackage{pifont}
\newcommand{\cmark}{\ding{51}}%
\newcommand{\xmark}{\ding{55}}%

\usepackage{makecell}
\usepackage{array}
\usepackage{enumitem}

\newcommand{\toprule}{\hline}
\newcommand{\midrule}{\hline}
\newcommand{\bottomrule}{\hline}

\makeatletter
 \let\old@ps@headings\ps@headings
 \let\old@ps@IEEEtitlepagestyle\ps@IEEEtitlepagestyle
 \def\confheader#1{%
 \def\ps@IEEEtitlepagestyle{%
 \old@ps@IEEEtitlepagestyle%
 \def\@oddhead{\strut\hfill#1\hfill\strut}%
 \def\@evenhead{\strut\hfill#1\hfill\strut}%
 }%
 \ps@headings%
 }
 \makeatother

\confheader{%
 \small{Accepted for publication in IEEE Transactions on Circuits and Systems for Artificial Intelligence. DOI: 10.1109/TCASAI.2025.3565685}
 }

\begin{document}
\bstctlcite{IEEEexample:BSTcontrol} 

\title{TransAxx: Efficient Transformers \\ with Approximate Computing}

\author{Dimitrios Danopoulos,
        Georgios Zervakis,
        Dimitrios Soudris,~\IEEEmembership{Member,~IEEE,}
        and J{\"o}rg Henkel,~\IEEEmembership{Fellow,~IEEE}

\thanks{Manuscript received September 30, 2024; revised March 10, 2025; accepted April 26, 2025.(\textit{Corresponding author: Dimitrios Danopoulos, e-mail:dimdano@microlab.ntua.gr}).}%
\thanks{D.~Danopoulos and D.~Soudris are with the School of Electrical and Computer Engineering, National Technical University of Athens, Athens 15780, Greece.}%
\thanks{G. Zervakis is with the Department of Computer Engineering \& Informatics, University of Patras, Patras, Greece}%
\thanks{J. Henkel is with the Chair for Embedded Systems at Karlsruhe Institute of Technology, Karlsruhe 76131, Germany.}%
\thanks{This work has been partially funded from the European Union's Horizon Europe research and innovation funding programme under grant agreement No 101070374 CONVOLVE (https://convolve.eu) and and by the German Research Foundation (DFG) through the project ``NA3OS: Neural Approximate Accelerator Architecture Optimization for DNN Inference on Lightweight FPGAs,'' HE 2343/22-1.}
}



\maketitle

\begin{abstract}
Vision Transformer (ViT) models which were recently introduced by the transformer architecture have shown to be very competitive and often become a popular alternative to Convolutional Neural Networks (CNNs).
However, the high computational requirements of these models limit their practical applicability especially on low-power devices.
Current state-of-the-art employs approximate multipliers to address the highly increased compute demands of DNN accelerators but no prior research has explored their use on ViT models.
In this work we propose TransAxx, a framework based on the popular PyTorch library that enables fast inherent support for approximate arithmetic to seamlessly evaluate the impact of approximate computing on DNNs such as ViT models.
Using TransAxx we analyze the sensitivity of transformer models on the ImageNet dataset to approximate multiplications and perform approximate-aware finetuning to regain accuracy.
Furthermore, we propose a methodology to generate approximate accelerators for ViT models. 
Our approach uses a Monte Carlo Tree Search (MCTS) algorithm to efficiently search the space of possible configurations using a hardware-driven hand-crafted policy.
Our evaluation demonstrates the efficacy of our methodology in achieving significant trade-offs between accuracy and power, resulting in substantial gains without compromising on performance.

\end{abstract}


\section{Introduction}
\label{sec:introduction}

Deep Learning (DL) has achieved remarkable success in a vast range of applications such as image processing, where it has emerged as one of the most powerful and accurate techniques. CNNs are capable of achieving high accuracy and performance on visual recognition and complex regression algorithms. In addition to CNNs, recent advancements in deep learning have led to the emergence of Vision Transformer (ViT) models, a new type of neural network architecture based on the self-attention mechanism of transformers which have achieved state-of-the-art performance on various computer vision tasks. However, ViT models are very computationally expensive due to their large number of parameters and the self-attention mechanism used in their architecture. Their high computational demands have limited their applicability on resource-constrained devices.

The utilization of approximate computing has demonstrated potential in enhancing the efficiency of deep learning models by reducing their computational complexity and memory demands \cite{DAN2021100520}.
It involves trading off a small amount of accuracy in DNNs for significant gains in speed and power efficiency by using inexact arithmetic components in place of their accurate counterparts~\cite{AxDNNsurvey,axxtrain, 10.1007/978-3-030-79025-7_25}.
Although, approximate computing has emerged as a promising approach to improve the efficiency of DNNs, no prior research has investigated its exploitation and applicability on transformers.

The wide-ranging and extensive domain of approximate compute units (ACUs) and their non trivial impact on the DNN accuracy increase the complexity of designing such hardware. Thus, the need for an approximate emulation framework that can address this intricacy becomes apparent.
Although several studies focus on the error metrics that should be optimized when designing approximate multipliers for DNN accelerators~\cite{adapt,tfapprox,proxsim,approxtrain}, predicting the exact impact of an approximate multiplier on the accuracy of a ViT model is highly challenging and cannot be analytically determined. Information distortion of the self-attention map of ViTs, which involves a large number of operations, may lead to large errors being introduced by lower precision \cite{li2022qvit}. 
Additionally, previous studies~\cite{adapt,tfapprox,weightoriented}, some of which employed GPU acceleration~\cite{proxsim, 10059158}, primarily focus on CNNs and not Transformers.
DNNs can often become sensitive to approximation~\cite{weightoriented} and hence, approximation-aware retraining is required to recover the error introduced by approximation~\cite{weightoriented,tfapprox,axconvar}.
In addition, determining the appropriate approximate multiplier for each DNN layer (cross-layer) is of utmost importance when targeting to maximize the power gains under accuracy loss constraints~\cite{tfapprox}.
Many works have examined automated methods for determining the optimal per-layer quantization in quantized DNNs \cite{8954415, Lou2019AutoQAK, Yao2021}.
However, few previous works \cite{alwann} have examined an automatic search flow in approximate DNNs. If ViT models were intended for use, especially for large datasets such as ImageNet, the computational cost would be tremendous, thus it would be undoubtedly a design challenge. 
For example, a standard design approach would require tools such as Tangram ~\cite{10.1145/370155.370339}, SCALE-Sim ~\cite{scalesim}, Maestro ~\cite{9076333} or HAT ~\cite{hat} for architectural exploration and EDA tools to assess the hardware characteristics (energy, area, power, latency) of the processing elements (PEs) and memory. The information from the EDA tools is then integrated into the architecture exploration~\cite{9076333} tools to enable respective design space exploration, such as memory size, number of PEs, type of PEs, dataflow (weight-stationary, input-stationary, etc.), to satisfy the desired power-energy-latency-area trade-off.

In this paper, we present TransAxx, a fast emulation framework of approximate multipliers in ViT models.
TransAxx is developed on top of PyTorch and can run on Nvidia GPUs.
The objective of TransAxx is to streamline and accelerate seamlessly, for the first time, the process of simulating popular ViT models on approximate hardware.
It acts as a seamless PyTorch plugin that can be enabled by the user without obstructing the natural flow of the DL framework. This is a novel emulation platform, enabling support for the widespread PyTorch framework whose ecosystem has been known for taking over AI researchers~\cite{stateai}. It can efficiently handle all popular ViT models and perform approximate inference supporting mixed approximation (per-layer) as well. Additionally, a calibrated post-training quantization and approximate re-training is also supported for further accuracy improvement.
Finally, we use TransAxx and present an algorithm for automatic per-layer approximation that efficiently traverses the tree of possible solutions. Specifically, we developed a Monte Carlo Tree Search (MCTS) based method (which is often used as component in RL tasks) along with a dynamic hardware-driven policy to reduce the search process. 

In brief, if a hardware designer has produced a deterministic approximate multiplier, they can use TransAxx to assess the accuracy of ViT models when run on accelerators using this approximate multiplier. Considering architectures such as~\cite{alwann,retsina,weightoriented, 8815818}, if the designer has developed a library of approximate multipliers or approximate reconfigurable multipliers, TransAxx combined with our MCTS search can find optimal trade-offs between accuracy and power (or other metrics if needed) by selecting the optimal approximate multiplier (or configuration) for each layer. 
In these cases, the approximation-aware ViT training feature of TransAxx will also help to potentially recover any accuracy loss. The bit-width selection for multipliers and their power, performance and area (PPA) values do not affect TransAxx but are rather parameters exploited in a hardware optimization/design space exploration. Overall, TransAxx is essential for exploring the accuracy-hardware savings space of approximate multipliers for ViT models. It allows for fast evaluation of the accuracy of such models with approximate circuits, thereby enabling efficient design space exploration and co-design approaches.
Our contributions can be summarized as follows:

\begin{enumerate}
  \item We propose TransAxx, the first framework to seamlessly and efficiently simulate approximate ViT models.
  \item We evaluate, for the first time, the impact of approximate multipliers on popular pre-trained ViT models on ImageNet.
  \item We use an optimized post-training quantization and approximate-aware finetuning strategy to recover the accuracy loss due to approximation.
  \item We propose a MCTS-based search method to flexibly balance the exploration and exploitation, while considering inference power reduction and model accuracy. 
  \newline
\end{enumerate}

The rest of the paper is organized as follows: Section 2 reviews related work on quantized ViT models and approximate multipliers and CNNs. Section 3 outlines TransAxx's operation and fast GPU-based ViT emulation. Section 4 details accuracy-power design space exploration of approximated ViTs using MCTS. Section 5 presents the evaluation setup and results. Section 6 provides a discussion on TransAxx framework, and Section 7 concludes with a summary of contributions in the software-hardware ecosystem..

\section{Related Work}
\label{sec:related}

Taking cue from the remarkable achievements of transformer models in the domain of natural language processing (NLP), scientists have recently utilized transformer models to address computer vision (CV) challenges. 
Furthermore, in the relevant scientific literature, approximate multipliers have been widely explored as a means to reduce the energy and power consumption of DNN accelerators~\cite{AxDNNsurvey}.
While the impact on hardware metrics such as energy, power, and frequency can be directly obtained using EDA tools by simply replacing the exact multiplier with an approximate one, assessing the implications for accuracy is significantly more complex.
This complexity increases with larger or more intricate DNNs, such as ViTs ~\cite{weightoriented}.
Moreover, it is crucial to evaluate the impact on accuracy before undertaking more time-consuming hardware design phases, such as synthesis and place-and-route of the entire accelerator/system.
Last, automated mixed precision techniques for quantized CNNs have been already investigated by the community to strike a balance between the computational efficiency and the numerical stability but there is no research for approximate ViT models.

\subsection{Quantized ViT models}
Li et al. proposed Q-ViT \cite{li2022qvit}, a fully differentiable quantization technique for ViT models that focuses on a head-wise bit-width and switchable scale. Ding et al. \cite{10.1145/3503161.3547826} proposed APQ-ViT which introduced a unified bottom-elimination blockwise calibration scheme to surpasses the typical post-training quantization which causes significant performance drops. Last, Liu et al. \cite{liu2021posttraining} presented an effective post-training quantization algorithm that finds the optimal low-bit
quantization intervals for weights and inputs and introduced a ranking loss to keep the relative order of the attention layer values.

\subsection{Approximate Multipliers}\label{subsec:axmults}
Hardware approximation techniques can be applied at three levels: algorithmic/architectural, logic, and circuit~\cite{Zervakis:ISLPED2015,crosslayer:dac2016,Zervakis:TCASII2019,Bahoo:TETC2023}.
At the algorithmic/architectural level, approximation techniques modify or simplify the system architecture or the implemented function~\cite{gear,perforation}.
At the logic level, these techniques simplify or manipulate the implemented logic, such as by removing or replacing gates or connections and simplifying component truth tables, e.g.,\cite{7926993, axcompr, gateprune, axagaing}.
At the circuit level, the two most common techniques are voltage over-scaling (VOS) and overclocking\cite{Zervakis:ISLPED2015,Zervakis:TCASII2019}, with VOS being more prevalent due to its significant power savings.

Research on approximate multipliers has attracted considerable interest, particularly because multipliers are the most area- and power-intensive components in typical processing elements (PEs) of DNN accelerators~\cite{AxDNNsurvey}.
The literature includes a vast range of approximate multipliers, encompassing algorithmically approximated~\cite{perforation}, logic-approximated~\cite{7926993}, and VOS-approximated~\cite{Bahoo:TETC2023} designs, as well as hybrid approaches combining multiple techniques~\cite{Zervakis:ISLPED2015, vader}.
Each of these approximations results in distinct error distributions and hardware savings, contributing to a rich design space.
This extensive variety of approximate multipliers underscores the need for tools like TransAxx, which enable rapid and seamless evaluation of their impact on the accuracy of complex DNNs such as the ViT models.
Additionally, this shows the significance of our MCTS-based method in guiding users through layer-to-approximation mapping.

TransAxx is designed to be method-agnostic regarding the underlying approximation and supports any arbitrary approximate multiplier, provided its output is deterministic and depends solely on the multiplicand values.
This constraint arises mainly from the need to enable fast approximation-aware ViT inference and re-training.
To achieve this, we model the behavior of approximate multipliers at a high level and integrate them in Python (see Section~\ref{subsec:implementation}).
As a result, TransAxx seamlessly supports logic- and algorithmic-approximate multipliers.
However, VOS-approximated multipliers are not inherently supported, as VOS errors stem from timing violations, causing the multiplier's output to depend on both the current and previous inputs~\cite{vossim}.

\subsection{Approximate CNN Circuit Design}
Approximate computing takes a different approach, on top of quantization, by intentionally introducing errors into computations in exchange for reduced computational cost. Several previous works have presented custom DNN frameworks to simulate the accuracy of approximate CNNs. For example, TFapprox \cite{tfapprox} or ProxSim \cite{proxsim}, are frameworks built on top of TensorFlow to emulate approximate circuits in CNNs using GPU accelerators. ALWANN \cite{alwann} presents a methodology to apply layer-wise approximation on 8bit ACUs with finetuning capabilities. The experimental results of these works, however, are limited mainly to ResNet CNN models for small datasets like Cifar-10. Other frameworks, such as AdaPT~\cite{adapt} and ApproxTrain~\cite{approxtrain} experimented on a wide range of DNNs but did not propose a design space exploration strategy while posing difficulties for the users to test their custom models. 

Several frameworks apply Design Space Exploration (DSE) by leveraging mixed precision techniques. Mixed precision techniques offer potential benefits in increasing computational speed and reducing model size. By exploiting model redundancy, these methods allocate lower bit-width datatypes to layers that are less sensitive or less critical, optimizing overall performance. However, the challenge lies in accurately measuring each layer's \textit{sensitivity} score and mapping it to the appropriate bit-width. A lot of techniques have been proposed to efficiently traverse this parameter space, such as HAWQV3 \cite{Yao2021}, a hardware-aware mixed-precision quantization formulation that uses Integer Linear Programming (ILP), or deep reinforcement learning (DRL) based quantization methods such as AutoQ \cite{Lou2019AutoQAK} and HAQ \cite{8954415}. While training RL agents for common hardware might be feasible, training such algorithms for simulated approximate CNNs, especially ViTs, would immensely exceed the computational and timing constraints. Approximate DNN frameworks run much slower than the default DL frameworks (i.e. \cite{tfapprox}, \cite{alwann}) due to the lack of adequate support for approximate arithmetic. 
Additionally, when assessed on new ACUs or models, the acquired policies of RL agents for determining optimal layer-wise approximations may deteriorate. This is due to the fact that each ACU may exhibit unique behaviors when encountering layers with varying data distributions. Consequently, the effectiveness of learned policies may diminish, hindering their transferability across different hardware configurations and model topologies.

\begin{figure*}[h]
\centering
  \includegraphics[width=0.95\linewidth]{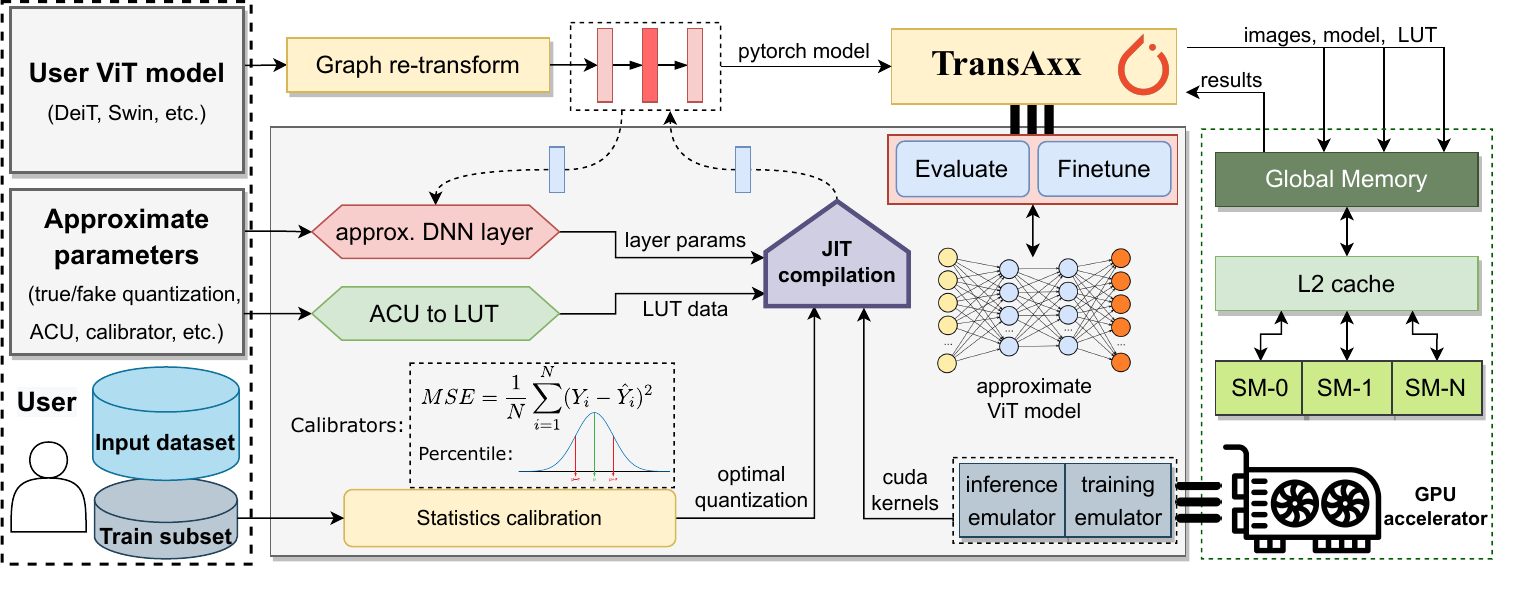}
  \caption{Abstract overview of TransAxx framework operation. The simulation flow starts from the user input on the left that comprises i) the ViT model in Pytorch and the respective train/test datasets, ii) the approximate multipliers, and iii) several other user-defined parameters required by TransAxx (quantization, calibration, etc). Then, TransAxx generates the LUTs that model the approximate multipliers and through the required transformations simulates the model's behavior under various conditions, such as different approximate multipliers per layer. Also, TransAxx facilitates fine-tuning through approximate-aware retraining.
Beneath the framework's simulation flow, a GPU is utilized to accelerate the process.}
  \label{fig:system}
\end{figure*}

\section{Fast Emulation of Approximate ViT Models}\label{sec:transaxx}

The TransAxx framework that we developed is intended for fast emulation of cross-layer DNN approximation and is available as a PyTorch plugin. This plugin can be activated or deactivated as desired by the user, allowing for the use of the PyTorch default flow when necessary. With TransAxx, a wide range of layer and model architectures can be seamlessly supported without any intervention by the user.
Though, in this work we focus on ViT models as approximations on such models have not been investigated yet.
Furthermore, we provide two key techniques for improving accuracy: post training quantization with state-of-the-art calibration and approximate-aware retraining.
Additionally, TransAxx supports mixed approximation (i.e. different multipliers) in between layers.
Finally, one of the primary challenges associated with using approximate components in DNNs is the need to emulate approximate operations fast, as existing DNN GPU-based accelerators do not inherently support such computations.
To tackle this issue, we implement a universal GPU accelerator which can run in all Nvidia GPU architectures in order to accelerate the emulation of approximate ViT models.

\subsection{Designing the framework}
\label{subsec:implementation}

Our framework, shown in \figurename~\ref{fig:system}, offers an orthogonal approach to simulating approximate ViT models. The primary functionalities are outlined below.

\begin{enumerate} 
  \item \textit{Extension of default PyTorch modules:} TransAxx aims to handle computations within approximate ViT models that involve non-differentiable operations or dependencies on non-PyTorch libraries. To achieve this, it extends the default PyTorch modules, allowing our custom functions to seamlessly integrate with the existing computational graph. During the model compilation, our framework automatically swaps the vanilla PyTorch layers with the custom layers, converting the default model to the desired approximate equivalent. These layers are instantiated on-the-fly using just-in-time (JIT) compilation, ensuring efficient integration with the model's computational graph. JIT also makes the model flexible and easy to modify during runtime. This method supports incremental compilation, which means that only the parts of the code that have changed are recompiled. This significantly reduces the overhead of repeatedly compiling and loading TransAxx's layer extensions during experimentation.
  
  \item \textit{Layer initialization and kernel dispatching:} Regular Tensor objects within PyTorch are leveraged to handle the initialization of weights or biases for the custom kernels. This ensures consistency with PyTorch's initialization mechanisms, maintaining compatibility and ease of use within the framework. Then the weights/activations are quantized based on the layer's multiplier bitwidth and calibrated using the statistics of the layer's activations. Last, our framework utilizes C++ macros to dispatch the appropriate GPU kernel per layer.

  \item \textit{Generation of Look-up Tables (LUTs):}   For each approximate multiplier, a corresponding LUT is generated from its high-level description (e.g., in C, Matlab, or behavioral HDL). We have an integrated tool within TransAxx that can generate this LUT for any arbitrary approximate multiplier, whose behavior is described in C or HDL. This is facilitated by running all possible hardware multiplications $x \times y$ (e.g., using an RTL simulation) for the given approximate multiplier and then storing the results in a LUT. Hence, \text{LUT}[$x$][$y$] gives the approximate product of $x$ and $y$. During the forward pass, TransAxx uses these LUTs and substitutes the default (exact) multiplication operator with the approximate product (i.e., loading the value \text{LUT}[$x$][$y$]). The LUT was a design choice to help reduce the emulation time of TransAxx. Initially, LUTs are stored in the global memory as it has a large size and it's the most appropriate option for the random access patterns of the LUTs. However, next, we show how we improved these memory accesses.
  We reiterate that TransAxx supports any arbitrary approximate multiplier as long as its output is deterministic and solely depends on the multiplicand values, without requiring the hardware approximate multiplier to be LUT-based.\label{rev:r2c1}
  

  \item \textit{GPU kernel optimization:} It is crucial to carefully consider memory transfers, as operations involving LUTs can quickly become memory-bound. Notably, the cache behavior of the GPU is influenced by both hardware specifications and the specific memory access patterns of the ViT model. However, given that LUT data is read-only, we can guide the Nvidia compiler to maximize memory access throughput (i.e., by using CUDA intrinsics and compiler flags). Using this approach, the LUT array will be typically cached through the L1 GPU cache, which offers low latency and can be shared among all threads within a CUDA core. This facilitates efficient caching and access to LUT data across multiple threads.
    
  \item \textit{Handling large bitwidths:}     
  For scenarios where LUTs may grow substantially in size, particularly with large bitwidths (\textgreater12 bits), our framework provides a flexible solution. TransAxx can dynamically substitute LUT-based multiplication with functional-based multiplication (in which the approximate multiplier is alternatively described in C-code). This process can introduce computational overhead in the DNN execution time but ensures that our framework remains efficient and scalable. It's worth mentioning that both approaches provide a 1-1 representation of the multiplier at high-level thus the results would be the same in inference or retraining. Also, it is important to note that transformers work well with low precision values~\cite{lin2023awq}, and higher bit-width, which might hinder TransAxx emulation time performance, is often not required.
  \end{enumerate}

\subsection{Support for the transformer architecture}

The transformer layer is the fundamental building block in the vision transformer architecture. Its purpose is to take a sequence of image patches as input, utilize the self-attention mechanism to establish long-range relationships, and produce a new feature sequence. There are also additional blocks at the beginning or at the end of a ViT model such as patch embedding or classification head. In patch embedding the input image is divided into fixed-size non-overlapping patches, and each patch is linearly embedded into a flat vector. These patch embeddings are then treated as the input tokens for the transformer model. At the end of the ViT model, there is typically a classification head responsible for making predictions based on the learned features. We chose to apply approximation only on the core transformer blocks involving the self-attention mechanism which largely dominate the execution time (usually $>98\%$). These blocks are often expanded to multiple transformer encoder blocks, each of which mainly contains a normalization, a multi-head self-attention, and a feed-forward layer. Towards incorporating approximate arithmetic, we will focus on the latter two which have the most mathematical operations.

The attention can be mathematically defined using the following equation, where the softmax function is used to compute the weights that determine the importance of each element in the input. Here, $Q$ is the query vector, $K$ is the key vector, and $V$ is the value vector:
\begin{align}
Attention(Q, K, V) = softmax(\frac{QK^T}{\sqrt{d_k}})V
\end{align}

In short, it calculates the association between the Query vector and the Key vector and multiplies the Value associated with each Key. Additionally, to maintain numerical stability and ensure that the attention scores have an appropriate variance, the similarity scores are scaled by dividing them by the square root of the dimensionality of the Key vectors, denoted as $d_k$.


\textbf{Multi-head Attention} Multi-head attention is a technique that uses multiple sets of (Q,K,V) triplets instead of just one set. This is done to account for cases where an element in a sequence has dependencies on more than one other elements. Using multiple weights associated with the same element provides a more comprehensive weighting of the sequence. The multi-head attention can be described as below.
\begin{gather}
\begin{aligned}
MultiHead(Q, K, V) = Concat(head_1, ..., head_h)W^O \\
\text{where} \quad head_i = Attention(Q W^Q_i, K W^K_i, V W^V_i).
\raisetag{22pt}
\end{aligned}
\end{gather}

Each head has its own set of learned parameters and performs the scaled dot-product attention operation separately.

\textbf{Feed-Forward Network} The Feed-Forward Network (FFN) is a two-layer classification network with a GELU (Gaussian Error Linear Unit) activation layer. It is designed to allow for non-linear interactions between patches or tokens in the input feature map, enabling the model to capture complex patterns and relationships. The FFN layer also involves a significant number of computations, so it is often necessary to implement approximate computing methods for this layer as well. It can be formulated as below:
\begin{align}
FFN(\vec{X}) = GeLU(\vec{X} W_1 + b_1) W_2 + b_2.
\end{align}
 where $W_1 \in \mathbb{R}^{d \times d_f}$, $b_1 \in \mathbb{R}^{d_f}$ and  $W_2 \in \mathbb{R}^{d_f \times d}$, $b_2 \in \mathbb{R}^{d}$. $d_f$ is the inner hidden size of the Feed-Forward Network.

\subsection{Quantization and fine-tuning strategies}

In order to simulate approximate computing units effectively, it is essential to implement an efficient quantization scheme that minimizes the impact of quantization errors. It is worth noting that many approximate multipliers, particularly those used in DNNs, are designed to support low bit-width integer (fixed-point) arithmetic \cite{AxDNNsurvey}. Previous studies on approximate CNN simulators have predominantly concentrated on standard 8-bit quantization \cite{tfapprox, proxsim}. However, in TransAxx, we employ a versatile bitwidth quantizer to enhance flexibility and accuracy. As our paper's scope does not involve proposing a new quantization method, we employed a state-of-the-art open-source quantizer based on Nvidia's TensorRT toolkit that can accommodate both lower and higher precisions. This flexibility can be crucial when simulating higher precision ACUs for deep neural networks that might lack error resilience \cite{quant1, quant2}. Also, it can allow researchers to understand the trade-offs between bitwidth, accuracy or computational efficiency.

The mapping between real and quantized values must be affine, defined by the equation  $Real\_Value = A \times Quantized\_Value + B$ where $A$ represents the scale factor and $B$ is typically set to zero. To determine optimal quantization parameters for the scale values, we employed a calibration technique to collect data statistics. Our quantization modules were equipped with a histogram calibrator aimed at achieving a 99.9\% percentile, which demonstrated superior overall performance.
 However, alternative methods like MSE (Mean Squared Error) or entropy can be applied transparently in our framework if desired. Furthermore, optionally after post-training quantization, we can perform Quantization Aware Training (QAT) by continuing training the calibrated model based on Straight Through Estimator (STE) derivative approximation. 
Notably, in Transaxx, QAT is approximation-aware as it simulates the approximate noise of ACUs during the finetuning stage and eventually shows more robustness to the applied multipliers. During approximate-aware retraining, TransAxx propagates the computations through our ACUs (usually for 2.5\% of the default training schedule) effectively computing the layer gradients using STE and increasing the final accuracy of the approximate ViT model at the end.  \label{par:quantization}

\begin{figure}[tb]
  \centering
  \includegraphics[width=0.9\columnwidth]{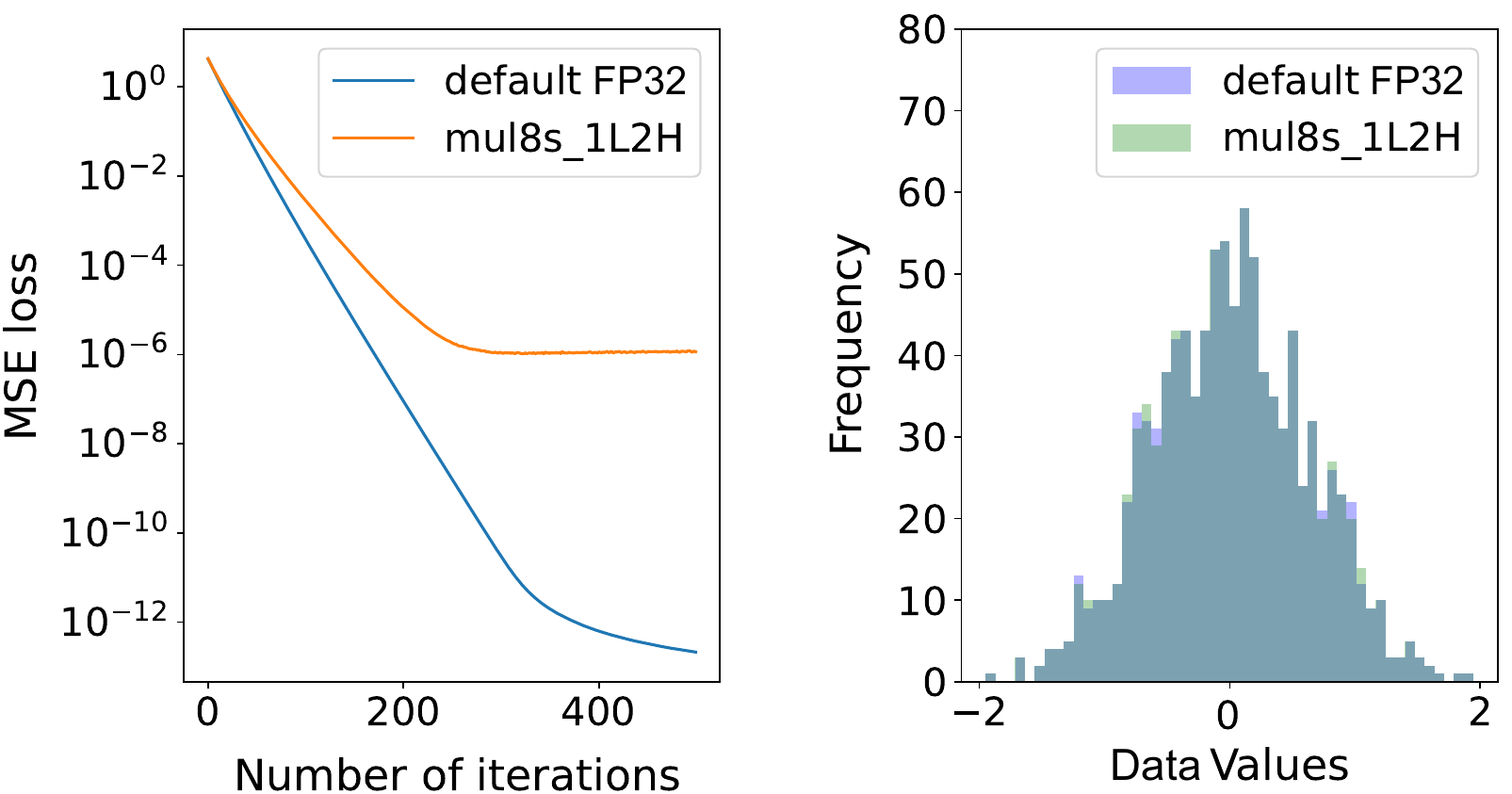}
  \caption{Preliminary testing with an approximate attention layer. \textbf{Left}: MSE loss per training iteration. \textbf{Right}: Histograms of target data (using FP32) and output data (using approx. multiplier) distributions from the layer's inference. }
  \label{fig:toy}
\end{figure}

\subsection{Toy experiment}

A preliminary toy experiment was conducted to test the effectiveness of forward/backward propagation on a simple approximate attention layer, using an 8-bit approximate multiplier instead of the default FP32 arithmetic of PyTorch. The core attention mechanism works by focusing on specific parts of the input sequence based on their relevance to the current output. The goal of this experiment is to see if a simple layer with the core attention mechanism can propagate correctly the data forward or backward and ensure, thus, that its functionality is not compromised by the approximate hardware emulation.

To perform the experiment, the attention layer was modified to use a \texttt{mul8s\_1L2H} approximate multiplier from the Evoapprox lib~\cite{7926993} for the forward and backward passes. The modified layer was then trained using a standard Stochastic Gradient Descent (SGD) algorithm for 500 iterations on data taken from $\mathcal{N}(0, 1)$. The results presented in \figurename~\ref{fig:toy} showed that backpropagation on our framework works as expected minimizing the MSE while the target values in the layer exhibit similarity with the accurate layer. Mathematically, the output data $Y$ converges in probability to the target data $X$ because as the number of samples (n) approaches infinity, the probability that the difference between $Y_n$ and $X$ being greater than some small value epsilon approaches zero.
\begin{align}
lim_{n->\infty} P(|Y_n - X| > epsilon) = 0.
\end{align}
Epsilon here is a threshold that specifies the maximum allowable error between the output and target values and it is ultimately dependent on the approximate multiplier used.

\section{Searching the Space of Approximate Designs}\label{sec:dse}

Exposing the optimal configuration of approximate multipliers between each layer of a DNN model in order to find the best trade-off between performance and power is liable to cause a significant computational overhead. The design space becomes large and measuring the accuracy of every configuration is not feasible even when using our GPU-based acceleration, particularly in our case, where the ViT simulation increases further the execution time. To efficiently explore the approximate design space we use a \textit{hardware-driven} MCTS to narrow down the architecture space for the approximate ViT models, maximizing accuracy while still meeting our given power constraints. To further reduce feedback time, we also developed an accuracy predictor for the inference accuracy. Our method reaches a near Pareto-optimal curve between power and accuracy.
The emphasis on power and accuracy reflects prevailing trends in the field where these metrics directly impact overall system performance and feasibility, especially under energy, power, and/or thermal constraints~\cite{AxDNNsurvey,axthermal}.
Nevertheless, the objective function used in our search is modular and can be extended to include other factors if needed.
While our current implementation does not explicitly consider latency and area, by modifying the objective function accordingly, such metrics can be incorporated in the analysis and design space exploration.\label{rev:r2c3}

\subsection{Rationale for employing Monte Carlo Tree Search}
\label{subsec:mcts_rationale}

MCTS is an AI search technique, often used in board games, that uses probabilistic and heuristic-driven algorithms to combine the classic implementation of tree search with principles from machine learning and particularly reinforcement learning. There has been only a few previous works that utilized MCTS on the AutoML domain, however it is a different domain than ours. Specifically some previous works have investigated the use of MCTS-based methods for hyperparameter tuning regarding CNNs.\cite{article1, Rakotoarison2019AutomatedML}. Also, regarding the design of electronic circuits there are also a few number of works but they focus on solving the routing problem. \cite{10.1115/1.4056221, 9768074, 10.1145/3505170.3506721}. Now, in our research area regarding approximate simulation frameworks, MCTS has not been investigated for finding optimal approximate configurations in DNNs, specifically ViTs. MCTS has the ability to dynamically balance exploration and exploitation, making it less susceptible to getting stuck in local optima compared with other methods, such as greedy algorithms, which are often used for quantization precision search \cite{greedycnn1, greedycnn2}. Specifically, the case of using inexact arithmetic can introduce additional sources of error that can make the optimization problem more complex to solve as it requires a greater degree of exploration than the greedy methods.

Furthermore, RL agents and genetic algorithms have been explored for optimizing hardware configurations in CNNs \cite{8954415}. However, applying these methods in our scenario is impractical for two primary reasons. Firstly, training and achieving convergence with an RL agent would demand an extensive amount of time, far exceeding our computational and timing constraints due to the need to simulate numerous ViT models. Additionally, the efficacy of learned agent policies in determining optimal approximations per layer could diminish when applied to new ACUs or models, as different ACUs exhibit substantial variation in behavior across layers with distinct data distributions.
Beyond RL approaches, when compared to genetic algorithms, MCTS offers the potential to efficiently and reliably identify effective solutions by balancing exploration and exploitation—a critical aspect in navigating this computationally intensive design space. Also, the effort required to optimize the parameters of the MCTS algorithm (e.g. exploration coefficient) is lower than the effort required for the genetic algorithms \cite{9466981}. Thus it would be problematic to apply genetic algorithms towards this complex problem which inhibits high variability across different ACUs. 

As we will discuss next, we propose a custom MCTS-based algorithm that is hardware-driven, meaning that for each action on the rollout phase it takes into consideration the sensitivity of the current DNN configuration to approximation. This hardware-driven MCTS method is another feature of our framework which enhances TransAxx by expanding its capability for design space exploration in ViTs.

\begin{algorithm}[tb]
  \caption{Pseudocode for our hw-driven MCTS}
  \label{alg:algorithm}
  \footnotesize
\textbf{Input:} 1) Model $M$, \ \ 2) Ground truth batch $B_g$, \ \ 3) Selected ACUs $A$, \\ 4) Exploration constant $c$, \ \  5) Rollout policy $P$, \ \  6) No. of Simulations $N$\\
\textbf{Output:} 1) Optimal Approx. Configs $C_{out}$
\begin{algorithmic}[1]
\State $rootNode \gets Node(M)$
\For{$i \gets 1$ to $N$}
\State \hspace{-0.1cm} $node \gets rootNode$
\While {\textbf{not} $node.isTerminal()$}
\If{$node.isFullyExpanded()$}
\State \hspace{-0.05cm} $node \gets node.getBestChild(c)$
\Else
\State \hspace{-0.05cm} $node \gets expand(node)$
\State \hspace{-0.05cm} \textbf{break}
\EndIf
\EndWhile
\State \hspace{-0.1cm} $state \gets node.state$
\While {\textbf{not} $state.isTerminal()$}
\State \hspace{-0.1cm} $a \gets chooseAction(state, P, A)$
\State \hspace{-0.1cm} $state \gets state.takeAction(a), a \in A$ 
\EndWhile
\State \hspace{-0.1cm} $Y_i \gets AxxConfig(state)$
\State \hspace{-0.1cm} $accuracy_i, power_i \gets evaluate(M, Y_i, B_g)$
\State \hspace{-0.1cm} $reward \gets accuracy_i - \lambda \times power_i$
\State \hspace{-0.1cm} $backprop(node, reward)$
\EndFor
\State $C_{out} \gets pareto(accuracy_i, power_i, Y_i) ,\forall i \in [1,N]$
\end{algorithmic}
\end{algorithm}

\subsection{The proposed algorithm}
\label{subsec:algorithm}
In MCTS, nodes are the building blocks of the search tree. The high level description of our approach (also shown as pseudocode in Algorithm \ref{alg:algorithm}) is as follows:

\begin{enumerate}
  \item Create a root node with an initial state of the model.
  \item Traverse the tree \textit{selecting} the node with the best Upper Confidence Bound (UCB) value according to~\eqref{eq:ucb} until a leaf node is reached.
  \item  If the leaf node is not terminal, \textit{expand} it by creating child nodes for all possible actions from that state.
  \item Simulate a \textit{rollout} from the selected child node based on the input policy $P$ until a terminal state is reached. Here, we take actions by choosing a potential ACU for the layer $l_i$ of the ViT model. We followed a head-wise approach similar to \cite{li2022qvit}, making decisions for individual heads within the multi-head attention layers of the ViT. The terminal state is defined as the point when all layers $l_i$ of the $L$ layers in the model have been assigned an ACU $A_i$. This can be expressed as ${Model(A_i)}_{i=1}^L$, indicating the model's configuration at the terminal state. Then, we compute the reward of the current approximate configuration of this rollout.
  \item \textit{Backpropagate} the reward obtained from the rollout up to root and update all UCB values of visited nodes.
\end{enumerate}

\begin{align}\label{eq:ucb}
S_i = x_i + c\sqrt{\frac{\ln N_i}{n_i}},
\end{align}
where $S_i$ the value of node $i$, $x_i$ the empirical mean of node $i$, $c$ the exploration constant, $N_i$ the total number of simulations up to $i$ and $n_i$ the number of visits of the node.

\captionsetup[sub]{font=footnotesize,labelfont={sf,sf}}
\begin{figure}[t]
\centering
\begin{subfigure}{0.24\textwidth}
\centering
\caption{VeiT-S}
\begin{tikzpicture}[scale=0.4] 
\begin{axis} [ybar,
ylabel={Normalized acc.},
    ylabel style={font=\Large},
    tick label style={font=\Large},
    label style={font=\Large},
    ymin=0,
    ymax=1.1,
    ytick={0.2,0.4,0.6,0.8,1}
]
\addplot [red!10!black,fill=red!60!black, opacity=0.7] coordinates {
    (1,0.587320141) 
    (2,0.901867468) 
    (3,0.941192842) 
    (4,0.956973087)
    (5,0.948067129)
};
\addplot [green!10!black,fill=cyan!60!black, opacity=0.7]  coordinates {
    (1,0.531456853) 
    (2,0.740311452)
    (3,0.830364747)
    (4,0.854905617)
    (5,0.87083747)
};

\end{axis}
\end{tikzpicture}
\end{subfigure}
\begin{subfigure}{0.24\textwidth}
\centering
\caption{DeiT-S}
\begin{tikzpicture} [scale=0.4] 
\begin{axis} [ybar,
    ylabel style={font=\Large},
    tick label style={font=\Large},
    label style={font=\Large},
    ymin=0,
    ymax=1.1,
    ytick={0.2,0.4,0.6,0.8,1},
]
\addplot [red!10!black,fill=red!60!black, opacity=0.7] coordinates {
    (1,0.541858882) 
    (2,0.864932315) 
    (3,0.682295999) 
    (4,0.657717613)
    (5,0.367743074)
};
\addplot [green!10!black,fill=cyan!60!black, opacity=0.7]  coordinates {
    (1,0.477266544) 
    (2,0.743844637) 
    (3,0.635198206) 
    (4,0.519369923)
    (5,0.391666529)
};
\end{axis}
\end{tikzpicture}
\end{subfigure}
\begin{subfigure}{0.24\textwidth}
\centering
\caption{Swin-S}
\begin{tikzpicture} [scale=0.4] 
\begin{axis} [ybar,
ylabel={Normalized acc.},
    ylabel style={font=\Large},
    xlabel={Layer},
    xlabel style={font=\Large},
    tick label style={font=\Large},
    label style={font=\Large},
    ymin=0,
    ymax=1.1,
    ytick={0.2,0.4,0.6,0.8,1}
]
\addplot [red!10!black,fill=red!60!black, opacity=0.7] coordinates {
    (1,0.990321398) 
    (2,0.998191372) 
    (3,0.990345839) 
    (4,0.995747281)
    (5,0.998631309)
};
\addplot [green!10!black,fill=cyan!60!black, opacity=0.7]  coordinates {
    (1,0.823093052) 
    (2,0.944967333)
    (3,0.902538944)
    (4,0.998849523)
    (5,0.998523508
)
};

\end{axis}
\end{tikzpicture}
\end{subfigure}
\begin{subfigure}{0.24\textwidth}
\centering
\caption{GCViT-XXT}
\begin{tikzpicture} [scale=0.4] 
\begin{axis} [ybar,
    ylabel style={font=\Large},
    xlabel={Layer},
    xlabel style={font=\Large},
    tick label style={font=\Large},
    label style={font=\Large},
    ymin=0,
    ymax=1.1,
    ytick={0.2,0.4,0.6,0.8,1}
]\addplot [red!10!black,fill=red!60!black, opacity=0.7] coordinates {
    (1,0.985096946) 
    (2,0.997617539) 
    (3,0.992751236) 
    (4,0.993993157)
    (5,0.998023064)
};
\addplot [green!10!black,fill=cyan!60!black, opacity=0.7]  coordinates {
    (1,0.861291085) 
    (2,0.954143857) 
    (3,0.877844115) 
    (4,0.882523174)
    (5,0.914506511)
};

\end{axis}
\end{tikzpicture}
\end{subfigure}
\caption{Comparison of actual (red) and predicted (blue) accuracy after applying approximation to each layer individually (from layer 1 to 5) across different ViT models.}
  \label{fig:head_axx}
  \vspace{-0.3cm}
\end{figure}

This iterative cycle of selection, expansion, simulation, and backpropagation continues until predetermined termination criteria are satisfied. At this juncture, the optimal action, typically the one leading to the most frequently visited state, is chosen. Each terminal state is represented as \( S = (A_{1}, A_{2}, \dots, A_{L}) \), encompassing all environment-specific information pertinent to the decision-making process, notably the power consumption of the current configuration and the output from the accuracy predictor. Additionally, the exploration-exploitation balance can be modulated using the variable \( c \) from the Upper Confidence Bound formula \eqref{eq:ucb}. Since evaluating the actual accuracy on the entire dataset (such as ImageNet in our case) is impractical, the predictor’s accuracy is computed during each Monte Carlo Tree Search (MCTS) rollout.
The Root Mean Square Error (RMSE) is employed to estimate the deviation between the predictor’s output and actual accuracy trends.
The predictor’s estimation is compared with the ground truth using the RMSE formula: \( \sqrt{\frac{1}{N}\sum_{i=1}^{N}(x_i - y_i)^2} \) for 128 input samples, which has been found sufficient for most scenarios. In general, determining the optimal placement of multipliers is challenging. The objective of MCTS is to explore alternative paths that might initially seem suboptimal to prevent the system from being trapped in local optima.
Overall, by employing a fast surrogate predictor in MCTS, we can navigate a much larger portion of the search space, leading to better overall design choices.
The RMSE accuracy predictor strikes a balance between speed and precision, enabling MCTS to efficiently converge toward near-optimal solutions by prioritizing a broad and efficient search.

\figurename~\ref{fig:head_axx} represents the normalized actual and predicted accuracy (red and blue bars respectively) after applying the \texttt{mul8s\_1L2H} ACU in the first 5 layers individually for every target model. Through these ablation studies we show our primary objective, which is not to achieve precise predictions of the accuracy, but rather to produce estimations that capture the general trend of the actual accuracy. Additionally, the figure illustrates that each layer block exhibits varying \textit{sensitivity} to accuracy perturbations, which is instrumental in refining the search strategy. This sensitivity information aids in guiding MCTS more effectively and prioritize more impactful configurations during the rollout phase, as discussed next.

\textbf{Defining a better policy:} 
The rollout policy often uses a straightforward heuristic to estimate the reward of a state by randomly selecting actions until reaching a terminal state \cite{randompolicymcts}. This is commonly done through a random policy, where actions are chosen uniformly at random, aiming to explore a wide range of states without any particular strategy. However, by incorporating domain-specific knowledge about the sensitivity of each layer to approximation, we devised a more advanced policy. Let $S = (s_{j,1}, s_{j,2}, \dots, s_{j,L})$ be the layer sensitivity list of an ACU $A_j$. $s_{j,i}$ is the normalized accuracy of the model when ACU $A_j$ with $j \in [1,k]$ is applied only on layer $i$. Similarly, we can represent the total returned power of the approximate model when $A_j$ is applied on layer $i$ as $p_{j,i}$. Conclusively, we can now express the probability of taking a specific action in the rollout policy, that is to select an $A_j$ out of $k$ available ACUs for layer $i$ as:
\begin{equation}\label{eq:policy}
P(A_j)_i = \frac{e^{(s_{j,i}- \lambda \times p_{j,i})}}{\sum_{z=1}^{k} e^{(s_{z,i}- \lambda \times p_{z,i})}}
\end{equation}

We incorporated expert knowledge to our problem by producing a better-informed rollout taking feedback from both sensitivity and power in each layer. Although we are making assumptions based on available knowledge, we incorporated randomness into the selection as well. This approach reflects \textit{optimism in the face of uncertainty} which as a mathematical concept has underpinned many decision-making algorithms~\cite{optimism1,optimism2}. In our case, this policy gave consistently stronger performance than the default random heuristic.

\section{Experimental Results}
\label{sec:exp_results}

The experiments in this study include the performance of TransAxx framework in terms of accuracy and execution time for popular ViT models and different approximate multipliers. Furthermore, we show the effectiveness of our MCTS algorithm for optimally searching the space of approximate designs to find the best balance between accuracy and power consumption. In terms of software versions, TransAxx was built on PyTorch 1.13 with CUDA Version 11.7. Also, the hardware setup we used was a 20-core Intel Xeon Gold 5218R server with 64GB of RAM and an Nvidia Tesla V100 GPU. 


\subsection{TransAxx performance on ViT models}
\label{subsec:performance}

Our framework was built with speed but also with flexibility in mind so that it can enable users to seamlessly test their custom ACUs fast. As aforementioned, two methods are supported during emulation; the LUT-based and the funtional-based approach. Table \ref{tab:speed_results} shows the inference time using the two approaches as well as the retraining time for an epoch for each model. The models used for our experiments are four popular ViTs, namely ViT \cite{dosovitskiy2021an}, DeiT \cite{pmlr-v139-touvron21a}, Swin \cite{Liu2021SwinTH} and GCViT \cite{hatamizadeh2022global}. All experiments are done on ImageNet2012 dataset with 128 batch size. Regarding the retraining scheme, we use Adam optimizer with a learning rate of $5e^{-5}$ for 2.5\% of the ImageNet train dataset. For Table \ref{tab:speed_results} metrics, the 8-bit \texttt{mul8s\_1KV9}~\cite{7926993} is used but the execution time is similar with any ACU of the same size. The LUT-based approach is usually not affected by the type of ACU (assuming same bitwidth) and is significantly faster than the functional-based approach, the latter of which only serves as a back-up method in TransAxx to alleviate any unforeseen memory issues. 

\begin{table}[t]
    \centering
  \caption{Emulation time in TransAxx for different ViTs}
  \label{tab:speed_results}
  \footnotesize
  \setlength{\tabcolsep}{2pt}
  \begin{tabular}{l|ll|llc}
    \toprule
    DNN & \makecell{ FLOPs }& \makecell{ Params} & \makecell{ Inference \\ (w/ func.)} & \makecell{ Inference \\ (w/ LUT)} & \makecell{ Retraining \\ (w/ LUT)} \\
    \midrule
    ViT-S   & 4.2G & 22.1M & 121 min & 6 min  & 5.5 min \\
    DeiT-S  & 4.2G  & 22.1M & 122 min & 6.1 min  & 5.8 min  \\
    Swin-S & 8.5G & 49.6M & 242 min & 13.1 min  & 13 min \\
    GCViT-XXT  & 1.9G & 12M & 43.5 min & 3 min  & 3.5 min \\

    \bottomrule
  \end{tabular}

\vspace{-1ex}
\end{table}

\begin{figure}[t]
  \centering
  \begin{subfigure}[b]{0.253\textwidth}
    \includegraphics[width=\textwidth]{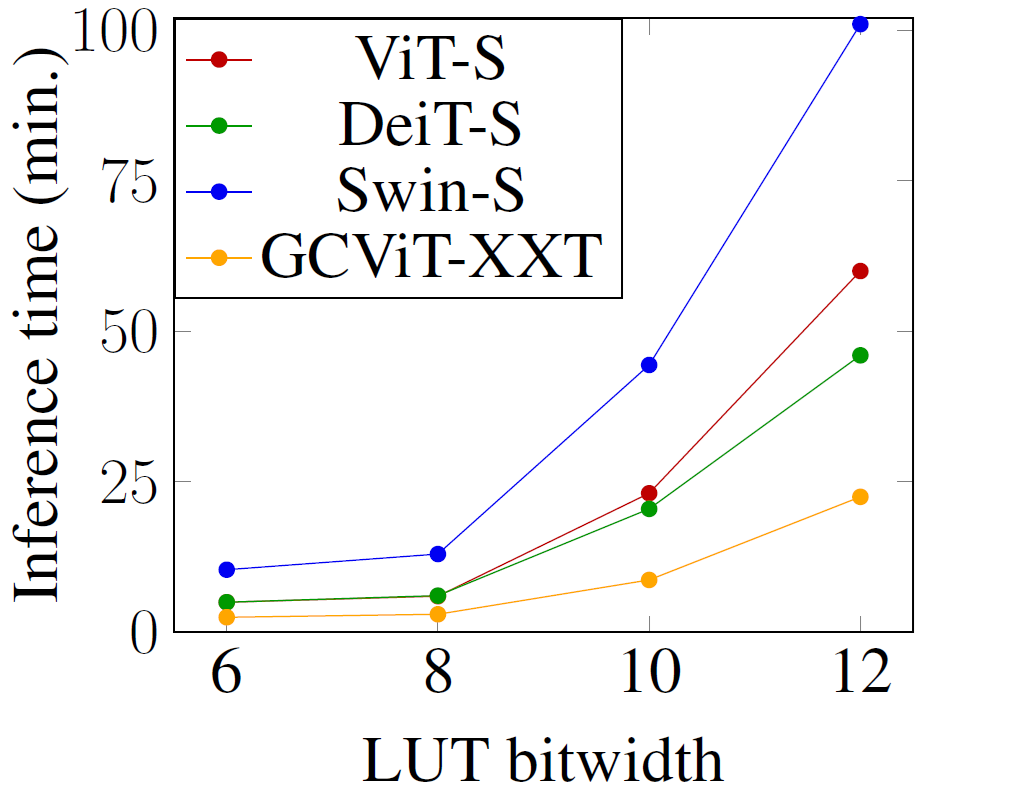}
  \end{subfigure}
  \hfill
  \begin{subfigure}[b]{0.227\textwidth}
    \includegraphics[width=\textwidth]{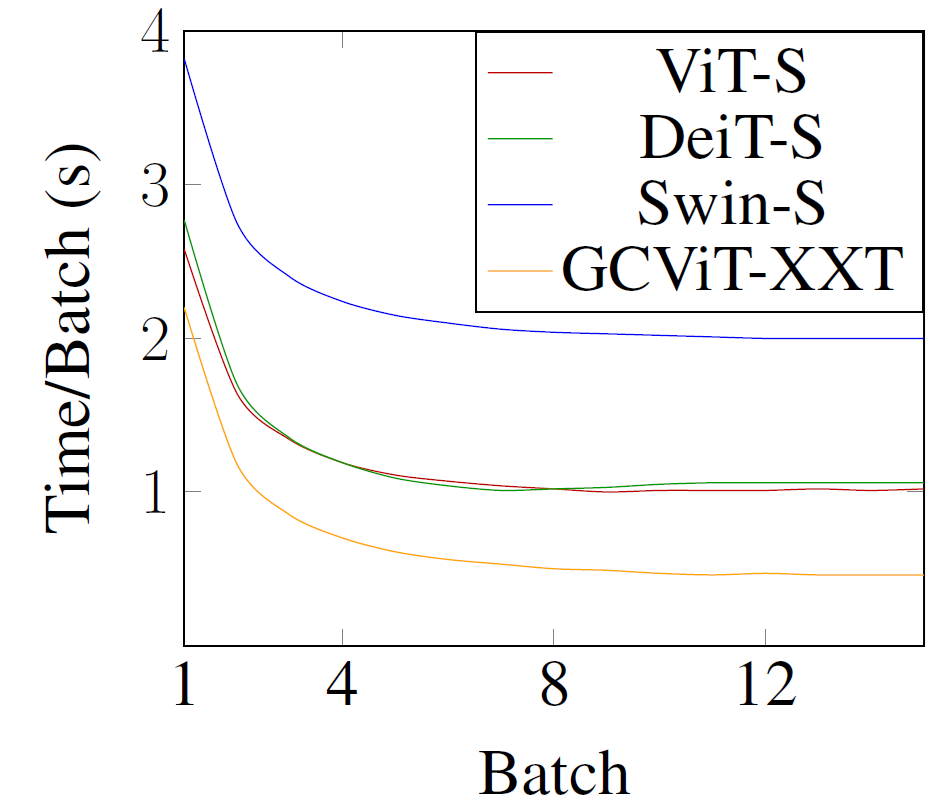}
  \end{subfigure}
  \caption{LUT-based multiplication performance. \textbf{Left}: LUT bitwidth impact on inference emulation time. \textbf{Right}: Caching effect on LUT performance during the first batches of inference emulation.}
  \label{fig:lut_mults}
\end{figure}

We further investigated the performance of LUT-based multiplication in our study, as depicted in \figurename~\ref{fig:lut_mults}. On the left side of the figure, we can observe the inference time for each ViT model using different bitwidths of the LUT. As it is evident, the emulatiom time increases as the memory requirements increase because it takes more time to fetch the LUT from the GPU memory. Also, larger LUTs may not fit entirely into cache, leading to increased cache misses and longer memory access times. In contrast, smaller LUTs are more likely to fit into cache, resulting in better cache utilization and lower memory access latency. Now, on the right side of the figure, we show the caching effect on the LUT performance across the first batches of the inference. As processing progresses through subsequent batches, the LUT caching mechanisms we introduced within TransAxx effectively come into play. These mechanisms allow for faster access to frequently used data, such as the LUT memory in our case, thereby reducing the computation time. As aforementioned, for \textgreater12 bits where LUT memory might increase substantially TransAxx can always subsitute the LUT-based with functional based approach. We report that for a 12-bit multiplier ( \texttt{mul12s\_2PP} from \cite{7926993}) the inference emulation time using its C functional description can be $\sim 5\times$ slower than the LUT-based approach.

In general, the execution times using the LUT-based approach are considered acceptable and fast enough, especially when considering that there is no other alternative for approximate ViT emulation.
Note also that these numbers regard the complex and large ImageNet dataset.

\begin{table}[t]
    \centering
  \caption{Qualitative comparison with state-of-the-art}
  \label{tab:tool_support}
  \footnotesize
  \setlength{\tabcolsep}{2pt}
  \renewcommand{\arraystretch}{1.1}
  \begin{threeparttable}
  \begin{tabular}{l|c|cccccc}
    \toprule
    
 Tool Support & TransAxx & \cite{adapt} & \cite{approxtrain} & \cite{tfapprox} & \cite{proxsim} & \cite{alwann} &  \cite{8714855}
 \\
    \midrule
     Framework & PyTorch & PyTorch & TF & TF & TF & TF & C++ \\
     Backend & GPU & CPU & GPU & GPU & GPU & CPU & CPU\\
     ViT model support & \cmark & \xmark & \xmark & \xmark & \xmark & \xmark & \xmark\\
     Automatic layer swapping & \cmark & \xmark & \xmark & \xmark & \xmark & \xmark & \xmark\\
     Quantization calibration  & \cmark & \cmark & \xmark & \xmark & \xmark & \cmark & \xmark \\
     HW-aware retraining & \cmark & \cmark & \cmark & \cmark & \cmark & \xmark & \cmark\\ 
     Design space exploration & \cmark & \xmark & \xmark & \xmark & \xmark & \cmark & \xmark\\ \hline
  \end{tabular}

\end{threeparttable}
\end{table}

Moreover, TransAxx is faster compared to similar frameworks that do not support ViT models though.
In order to compare with prior work, we can focus our comparison on the execution time of ViT-S which is closely related to ResNet50 in terms of FLOPs (4.2G vs 3.87G). TransAxx achieves faster inference time compared with other GPU simulation frameworks such as ProxSim \cite{proxsim} (6min vs 107min) and ApproxTrain \cite{approxtrain} (6min vs 10.4min). It is important to highlight some key implementation differences between TransAxx and other state-of-the-art frameworks, which could account for the observed performance variations. TransAxx starts the quantization process at the PyTorch level. This means that the values are quantized (converted to lower-precision formats like int8) before they are sent to the GPU. Consequently, when TransAxx needs to perform operations on these values, it deals directly with already quantized int8 tensors. For example, ApproxTrain \cite{approxtrain} performs quantization within the CUDA kernel. This means that each floating-point value must be converted to an integer inside the GPU kernel during computation. TransAxx can communicate using int8 tensors between the CPU and GPU memory. This communication is more efficient as it involves smaller, lower-precision data, which reduces the transfer overhead. Other frameworks often have to handle floating-point to integer conversion within the CUDA kernel, adding extra computational overhead and potentially causing delays due to less efficient data handling and increased complexity inside the GPU.

Last, it's worth noticing Table \ref{tab:tool_support}, which shows the robustness of TransAxx in terms of features compared with the state-of-the-art, combining all the important functionalities. Furthermore, the ease of use and user-friendly interface of TransAxx contribute significantly to its robustness. The tool is designed to be intuitive, allowing users to access its comprehensive features without a steep learning curve.

\begin{table}[t]
    \centering
  \caption{Hardware and error characteristics of the considered approximate multipliers from~\cite{7926993}.}
  \label{tab:acu_specs}
  \footnotesize
  \setlength{\tabcolsep}{3pt}
  \begin{tabular}{l|ccc|c|c|c}
    \toprule
    Multiplier & MAE & WCE & MRE & power & area & delay \\
    \midrule
    mul8s\_1KV6  & 0.0\% & 0.0\% & 0.0\% & 0.425mW & 729.8um\textsuperscript{2} & 1.48ns \\
    mul8s\_1KV9  & 0.0064\% & 0.026\% & 0.90\% & 0.410mW & 685.2um\textsuperscript{2} & 1.47ns \\
    mul8s\_1L2H & 0.081\% & 0.39\% & 4.41\% & 0.301mW & 558.8um\textsuperscript{2} & 1.36ns \\
    mul8s\_1L2L  & 0.23\% & 1.16\% & 12.26\% & 0.200mW & 411.6um\textsuperscript{2} & 1.14ns \\

    \bottomrule
  \end{tabular}
  \end{table}

\subsection{Evaluation of Approximate Multiplication for ViT Models}
\label{subsec:mult_evaluation}
In this subsection we perform a systematic analysis on the top-1 accuracy achieved on ImageNet-1K dataset for each of the four aforementioned ViT models on the default FP32, 8-bit quantized, approximate and retrained versions. Apart from performing successful finetuning after approximation, we considered significantly the quantization part, as also described in Section \ref{par:quantization}. In particular, prior to the approximation of the models, we performed a calibrated quantization scheme since finding a scale parameter correctly is known to have a large impact on the network’s performance\cite{nagel2021white}.

\begin{table*}[t]
\centering
\caption{Accuracy and power benchmark [\%] per multiplier and model}
\label{tab:acc_results}
\footnotesize
\setlength{\tabcolsep}{5pt}
\begin{tabular}{|c|c|c|c||c|c|c||c|c|c||c|c|c|}
\hline
\multicolumn{4}{|c||}{Model specifications} & \multicolumn{3}{c||}{\makecell{Multiplier 1: \textbf{mul8s\_1KV9}}} & \multicolumn{3}{c||}{\makecell{Multiplier 2: \textbf{mul8s\_1L2H}}} & \multicolumn{3}{c|}{\makecell{Multiplier 3: \textbf{mul8s\_1L2L}}} \\
\hline
Model &\makecell{MACs \\ approx. \\ {[\%]}}  & \makecell{FP32 \\ {[\%]}} & \makecell{8bit \\ (calib.) \\ {[\%]}} & \makecell{Initial \\ {[\%]}}  & \makecell{Retrained \\ {[\%]}} & \makecell{Power \\ reduction \\ {[\%]}} $\downarrow$ &  \makecell{Initial \\ {[\%]}} & \makecell{Retrained \\ {[\%]}}  & \makecell{Power \\ reduction \\ {[\%]}} $\downarrow$ &  \makecell{Initial \\ {[\%]}} & \makecell{Retrained \\ {[\%]}}  & \makecell{Power \\ reduction \\ {[\%]}} $\downarrow$  \\
\hline
ViT-S & 98.54  & 74.64 & 71.86 & 34.95 & 67.31 & 3.45 & 1.264 & 66.74 & 28.75 & 0.090 & 0.15 &  52.18 \\
\hline
DeiT-S & 98.54 & 81.34 & 79.34 & 0.96 & 70.16 & 3.45 & 0.10 & 67.01 & 28.75 & 0.10 & 0.11 & 52.18  \\
\hline
Swin-S & 99.7 &  82.89 & 81.83 & 79.56 & 79.25 & 3.49 & 64.30 & 76.64 & 29.09  & 0.41 & 67.87 & 52.79   \\
\hline
GCViT-XXT & 75.5 & 79.72 & 78.91 & 73.50 & 78.346 & 2.64 & 51.56 & 76.93 & 22.03 & 0.26 & 63.01 & 39.98  \\
\hline
\end{tabular}
\end{table*}

Regarding the approximate multipliers used for the experiments, we chose four distinct multipliers with different error, power, area and delay characteristics, as summarized in Table~\ref{tab:acu_specs}. The ACUs were obtained from the open-source EvoApprox library~\cite{7926993} (\texttt{mul8s\_1KV6, mul8s\_1KV9, mul8s\_1L2H, mul8s\_1L2L}). Among these, \texttt{mul8s\_1KV6} was utilized as the accurate multiplier for the experiments, particularly in the context of 8-bit quantized models. All values regarding the specifications of each multiplier were obtained using Synopsys DC (45 nm PDK, 1 V, 25 ℃). We followed a fairly common approach used by state-of-the-art approximate DNN works~\cite{adapt, approxtrain, tfapprox, proxsim, alwann, 8714855, AxDNNsurvey} which also obtain their ACUs from EDA tools. Table~\ref{tab:acc_results} summarizes the accuracy results in percentages of the top-1 accuracy obtained before (Initial) and after (Retrain) applying approximate-aware retraining. Generally, we see that approximate-aware retraining reduces the accuracy gap successfully on the majority of approximate models, as the weights of the network can adapt to the distributions the ACUs represent. Additionally, we report the MAC (multiply-and-accumulate) power reduction. Clearly the actual power reduction would be influenced by numerous factors but the reduction from MAC operations usually has a cascading effect on the total consumption. Specifically, we compute the overall power reduction from the percentage of approximated MACs relative to the total MACs in the model. This means that the power savings are assessed based on how many of the total MAC operations have been approximated, highlighting the impact of approximated MACs on the model's power efficiency. As baseline we use the power consumption of the accurate \texttt{mul8s\_1KV6} multiplier. Accordingly, we can calculate the estimated MAC latency and area reduction for the multipliers \texttt{mul8s\_1KV9, mul8s\_1L2H, mul8s\_1L2L}, and we get a decrease of $0.47\%, 7.5\%, 21.4\%$ for delay and a decrease of $5.68\%, 21.8\%, 40.6\%$ for area respectively (the results for each multiplier are averaged from all four models).

\subsection{Exploring the design space}
\label{subsec:dse}

The manual and simple method of performing approximation, that is to apply the same multiplier to all layers in the model gave us substantial power gains with the cost of some accuracy drop as seen from Table \ref{tab:acc_results}. However, in some cases, it is not possible to recover the large impact that approximate multiplications had on accuracy.
Moreover, in many cases there might be a more power efficient solution that achieves similar accuracy. In this subsection, we conduct a comprehensive analysis of our automated search algorithm based on MCTS. The results of our automated search reveal a nuanced understanding of the trade-offs inherent in the power-accuracy space. We show that we can automatically find better solutions that are closer to the Pareto-front and thus offer better trade-off between power consumption and accuracy. For our experiment the four ACUs (\texttt{mul8s\_1KV9, mul8s\_1L2H, mul8s\_1L2L, mul8s\_1KV6}) are considered as possible candidates for each layer of every target model.

\begin{figure}
  \centering
    \caption*{\footnotesize \ \ \ \ \ \  \textcolor{orange!70!yellow}{\textbullet} mul8s\_1KV6 \ \ 
    \textcolor{blue!95!black}{\textbullet} mul8s\_1KV9 \ \ 
    \textcolor{green!60!black}{\textbullet} mul8s\_1L2H \ \
    \textcolor{red!75!black}{\textbullet} mul8s\_1L2L}
  \includegraphics[width=0.24\textwidth]{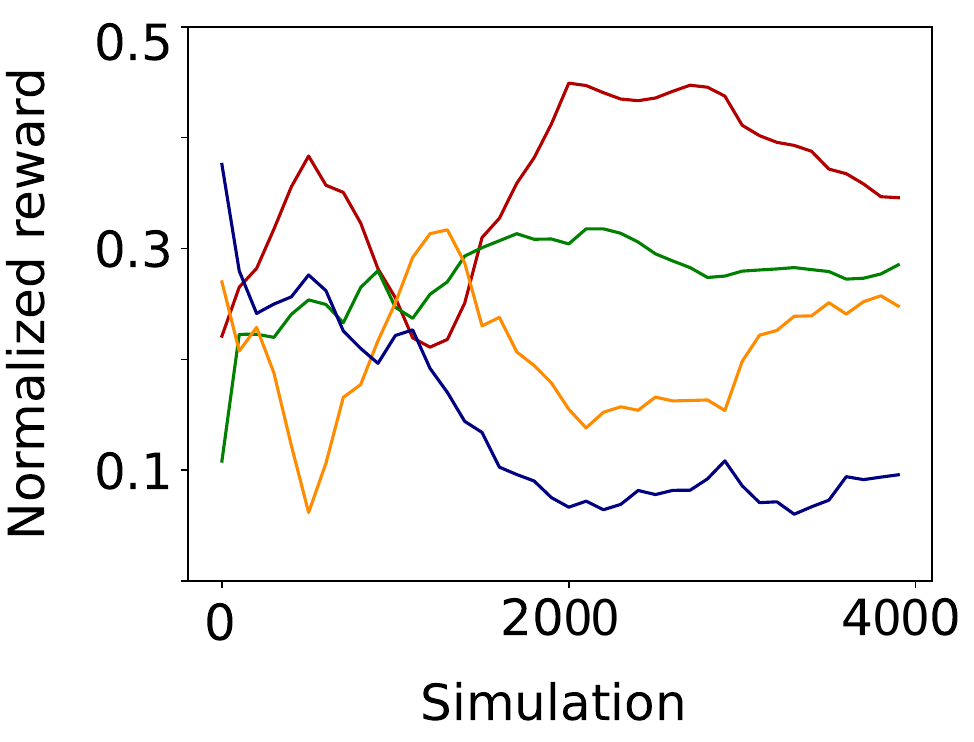}
  \hfill
  \includegraphics[width=0.24\textwidth]{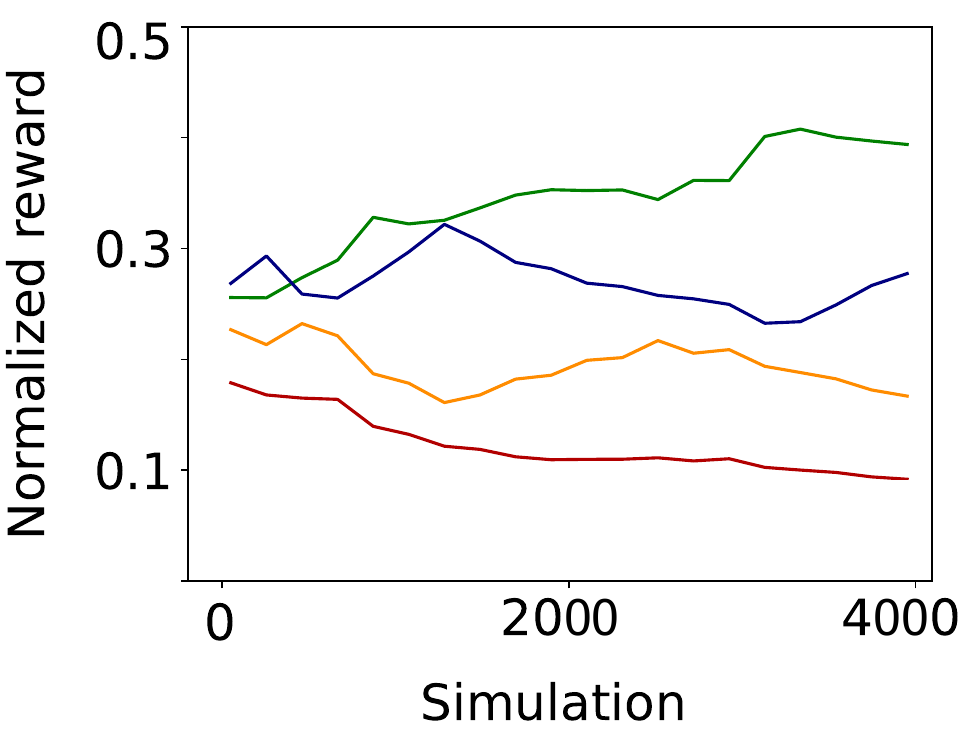}
  \caption{Rewards of each action from the MCTS root node using the random policy (left) and hw-driven policy (right).}
  \label{fig:mcts_actions}
  \vspace{-4ex}  
\end{figure}

\textbf{Policy evaluation and convergence}: Before proceeding with the Monte Carlo simulations it is essential to exhibit the stability of the algorithm and its ability to converge after some iterations. Naturally, our agent's performance is optimal when the rollout policy we used to estimate the expected rewards of each possible action is more likely to choose the best action. This is the reason we injected knowledge from the hardware configurations into the system using \eqref{eq:policy} so as to guide the search process towards the Pareto-optimal points. In \figurename~\ref{fig:mcts_actions}, for the case of ViT-S model, we measure the normalized reward for each of the four possible starting paths/actions from the root node of the MCTS tree. In each simulation, these rewards represent the estimated value of choosing a specific approximate multiplier based on the information gathered from the search tree. In this way, we demonstrate the ability of our custom hardware-driven policy to converge to more stable rewards faster than the random policy. Another valuable insight we can obtain from this figure is that the hardware-driven policy manages to find faster what might be a good or bad action to take according to the algorithm. For example, it prefers choosing \texttt{mul8s\_1L2H} multiplier at the first layer as it might give an "appealing" accuracy-power tradeoff which is also indicated by our measurements in Table \ref{tab:acc_results}. In contrary, the lower power \texttt{mul8s\_1L2L} multiplier is not often preferred as it significantly compromises the accuracy. We should note however, that intuitively these outcomes would vary across models, layers or multipliers. 

\begin{figure}
  \centering
  \includegraphics[width=\linewidth]{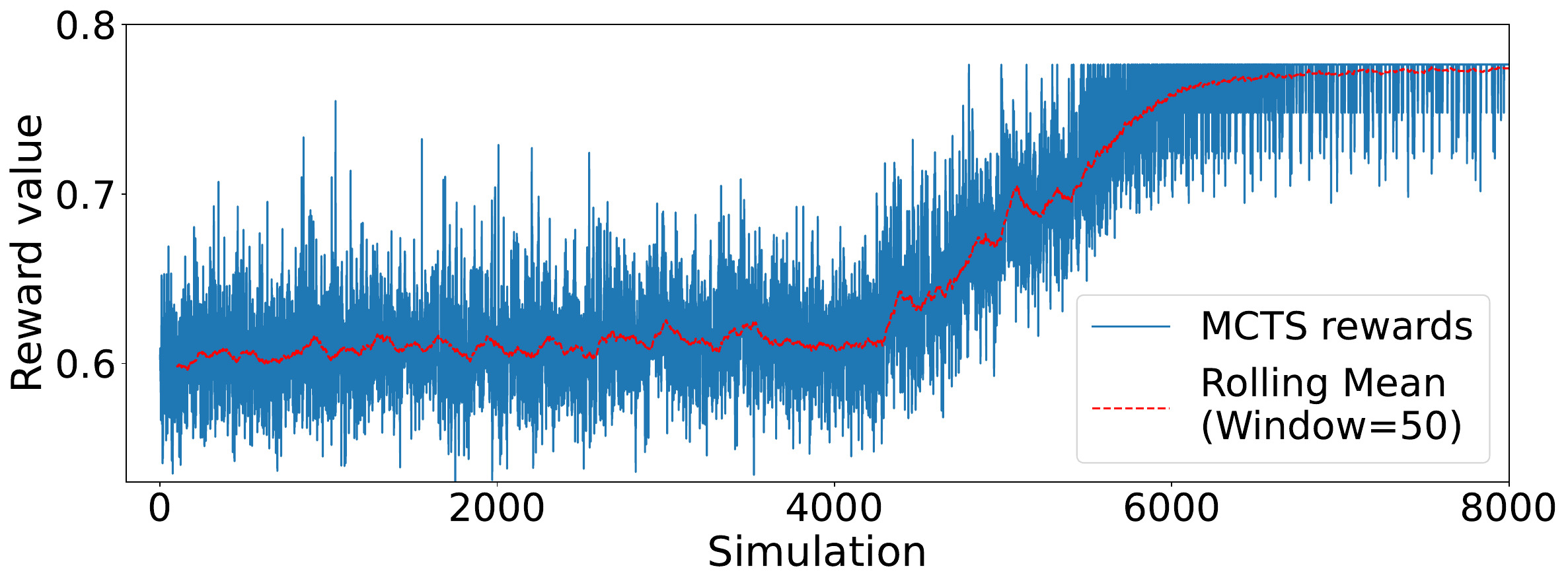}
  \caption{Convergence of the MCTS rewards.}
  \label{fig:mcts_convergence}
\end{figure}

Additionally, to demonstrate the convergence of our hw-driven MCTS-based search, we plot the reward values and their rolling mean (with a window of 50) targeting ViT-S model, over multiple simulations, as shown in  \figurename~\ref{fig:mcts_convergence}. As more simulations are performed, the MCTS tree is refined, and the algorithm converges towards optimal solutions. With each simulation, the search algorithm converges towards the approximate configurations that yield the best performance for the ViT model in terms of accuracy-power. The average reward stabilizes or reaches a plateau as the number of simulations increases.
This indicates that our algorithm explored the search space sufficiently and identified a solution (or set of solutions) that consistently achieves a certain performance level.


\begin{figure*}[ht]
    \centering
    \begin{subfigure}{0.24\textwidth}
        \centering
        \includegraphics[width=\linewidth]{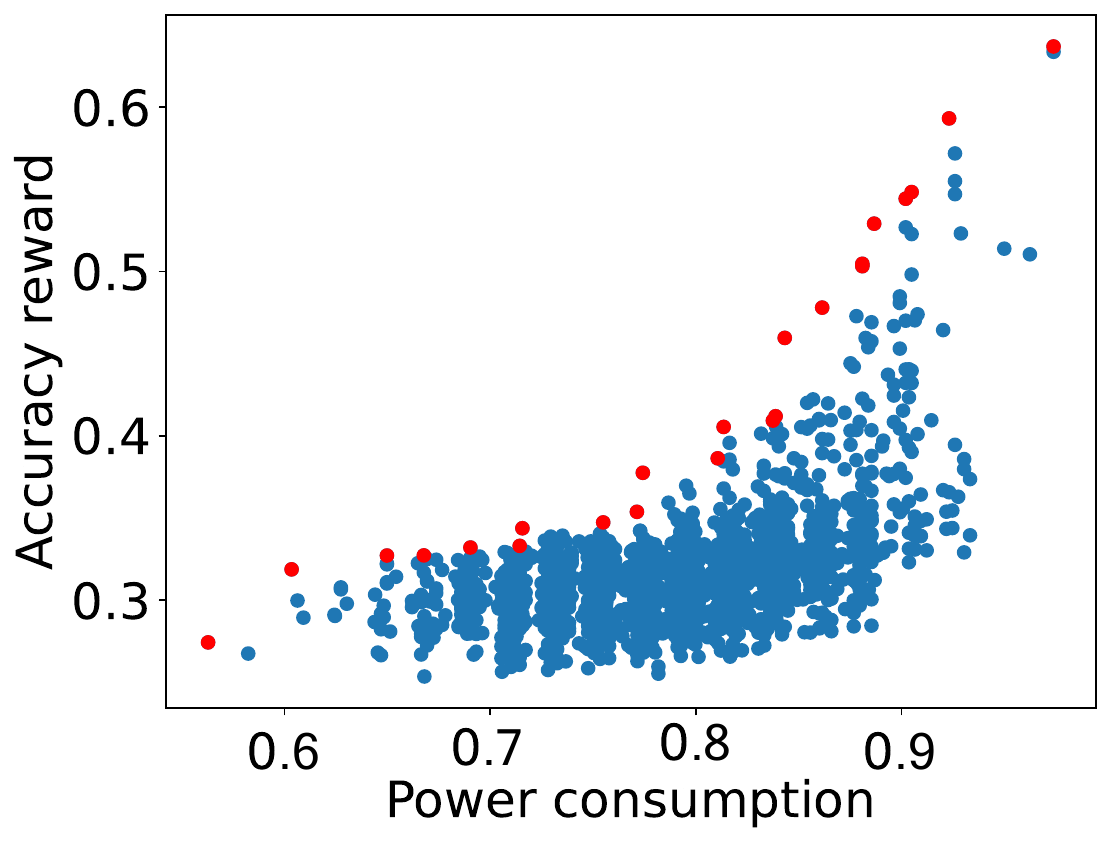}
        \caption{ViT-S}
    \end{subfigure}
    \hfill
    \begin{subfigure}{0.24\textwidth}
        \centering
        \includegraphics[width=\linewidth]{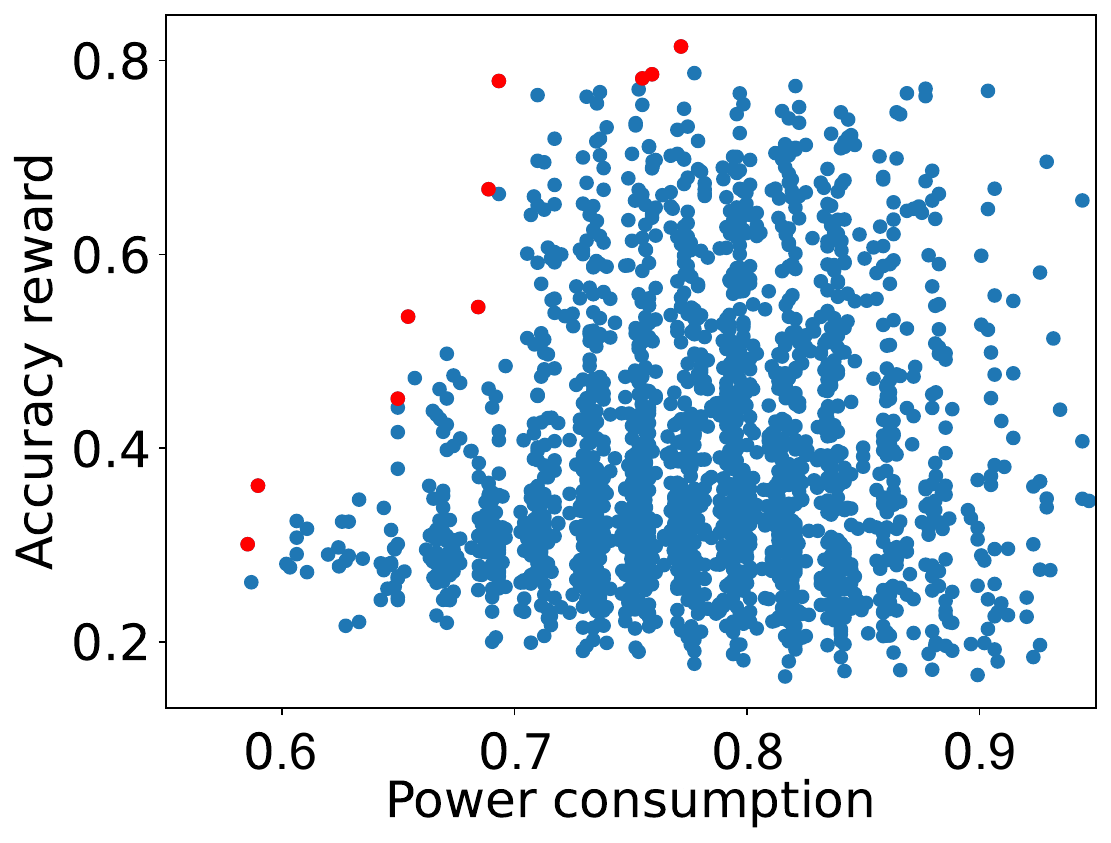}
        \caption{DeiT-S}
    \end{subfigure}
    \hfill
    \begin{subfigure}{0.24\textwidth}
        \centering
        \includegraphics[width=\linewidth]{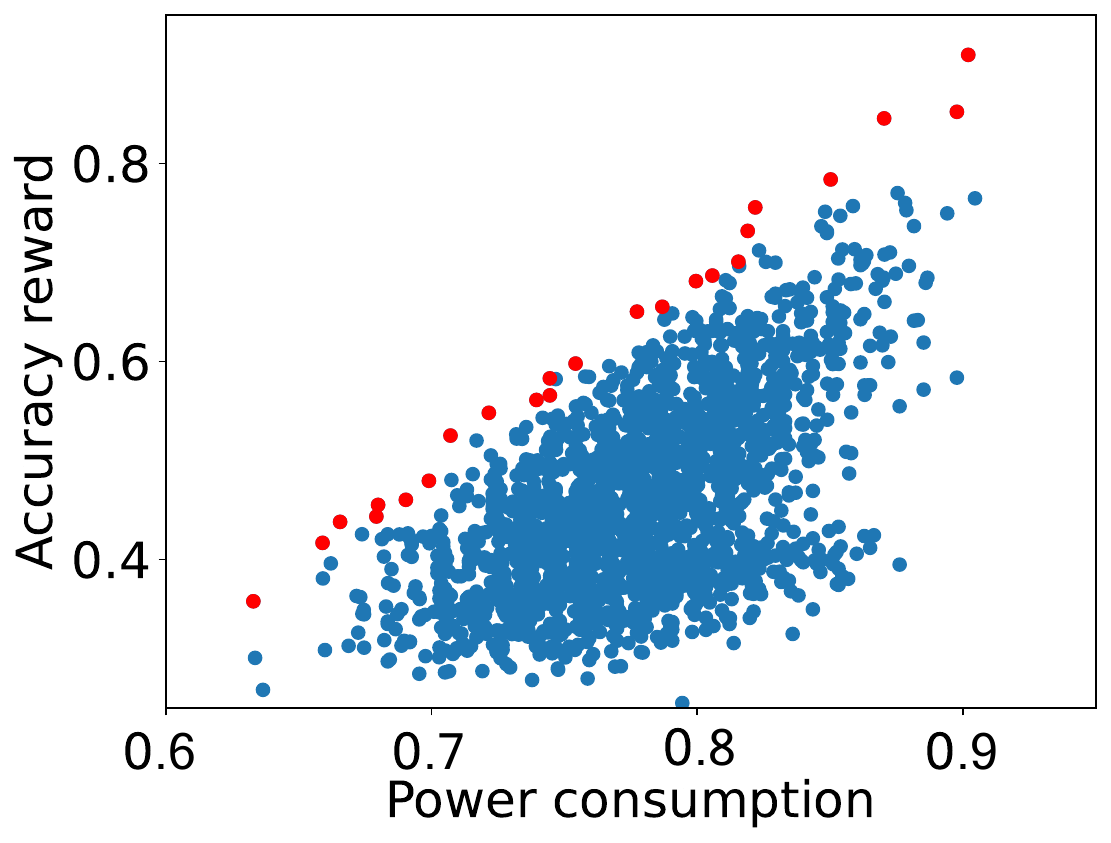}
        \caption{Swin-S}
    \end{subfigure}
    \hfill
    \begin{subfigure}{0.24\textwidth}
        \centering
        \includegraphics[width=\linewidth]{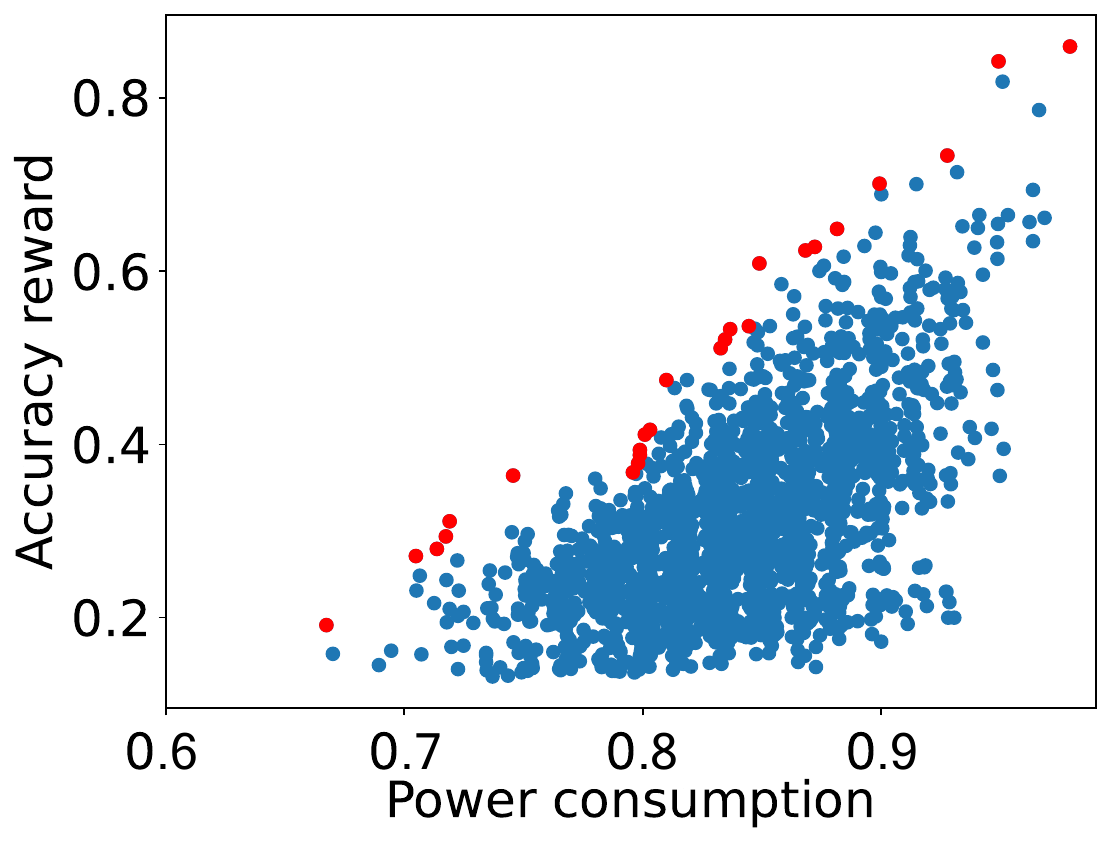}
        \caption{GCViT-XXT}
    \end{subfigure}
    
   \vspace{0.5cm}
   
       \begin{subfigure}{0.24\textwidth}
        \centering
        \includegraphics[width=\linewidth]{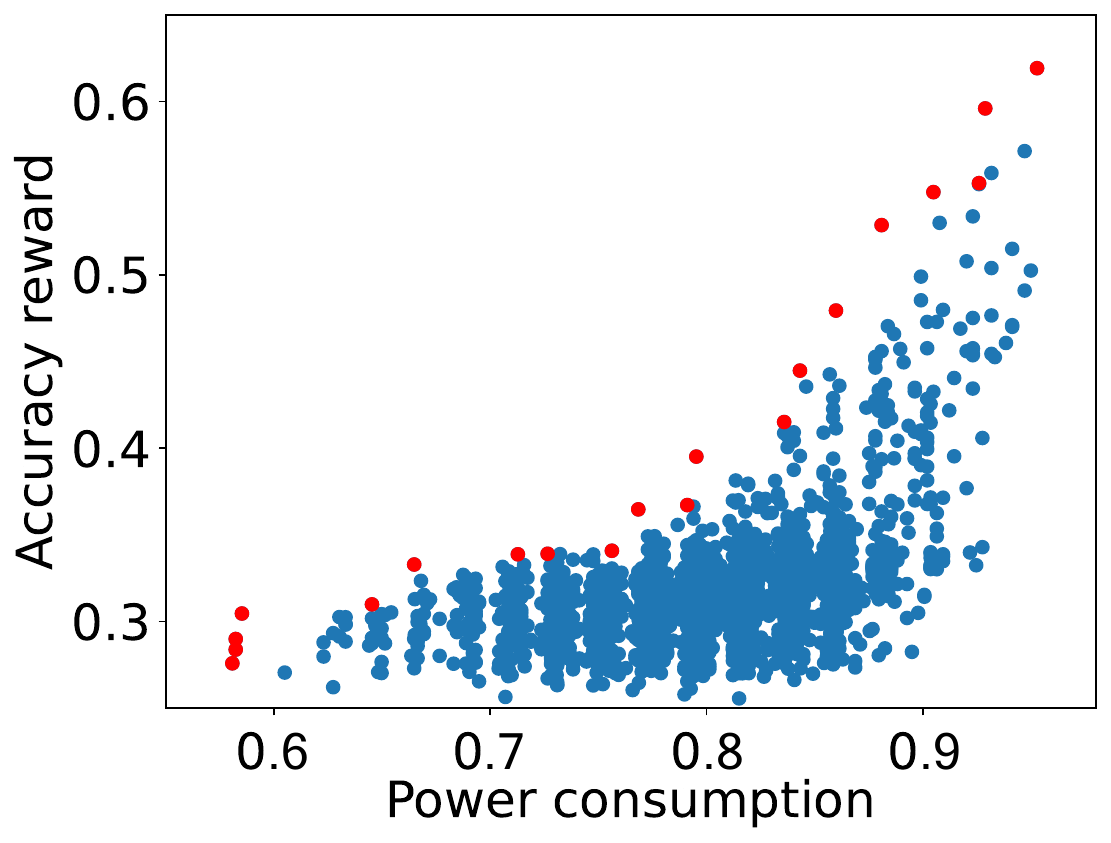}
        \caption{ViT-S}
    \end{subfigure}
    \hfill
    \begin{subfigure}{0.24\textwidth}
        \centering
        \includegraphics[width=\linewidth]{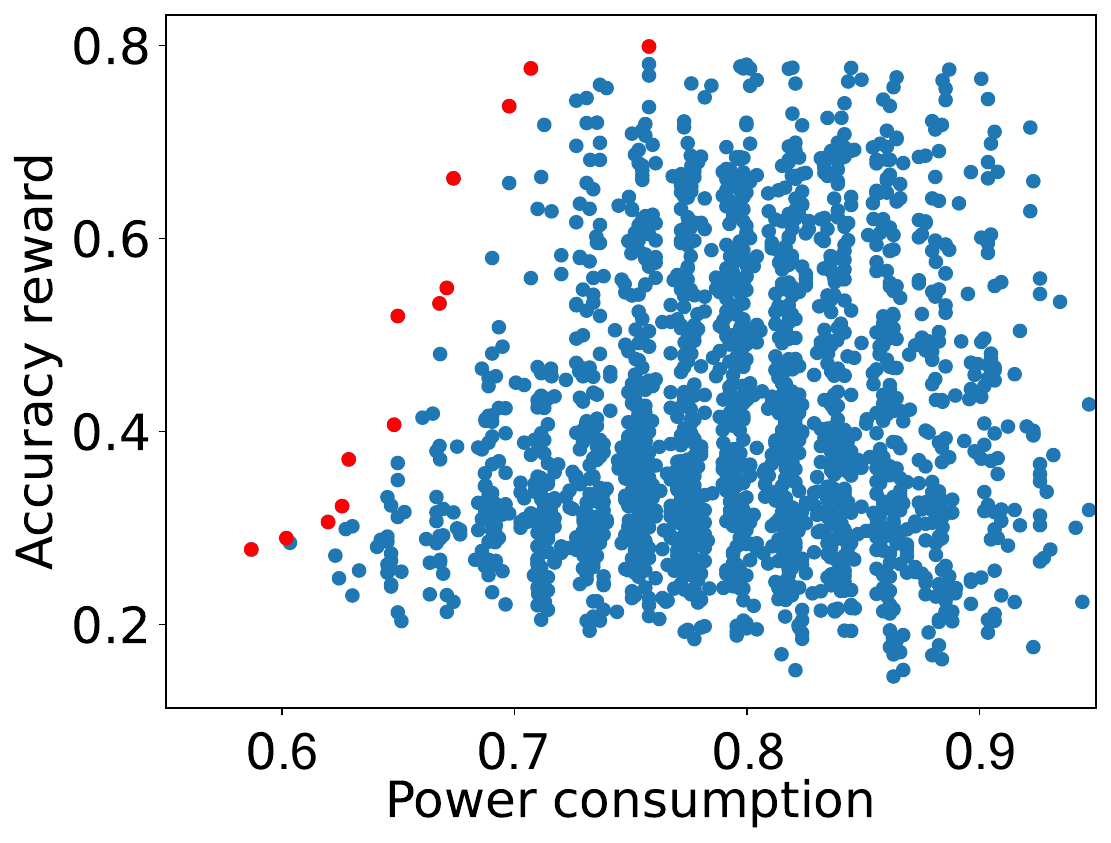}
        \caption{DeiT-S}
    \end{subfigure}
    \hfill
    \begin{subfigure}{0.24\textwidth}
        \centering
        \includegraphics[width=\linewidth]{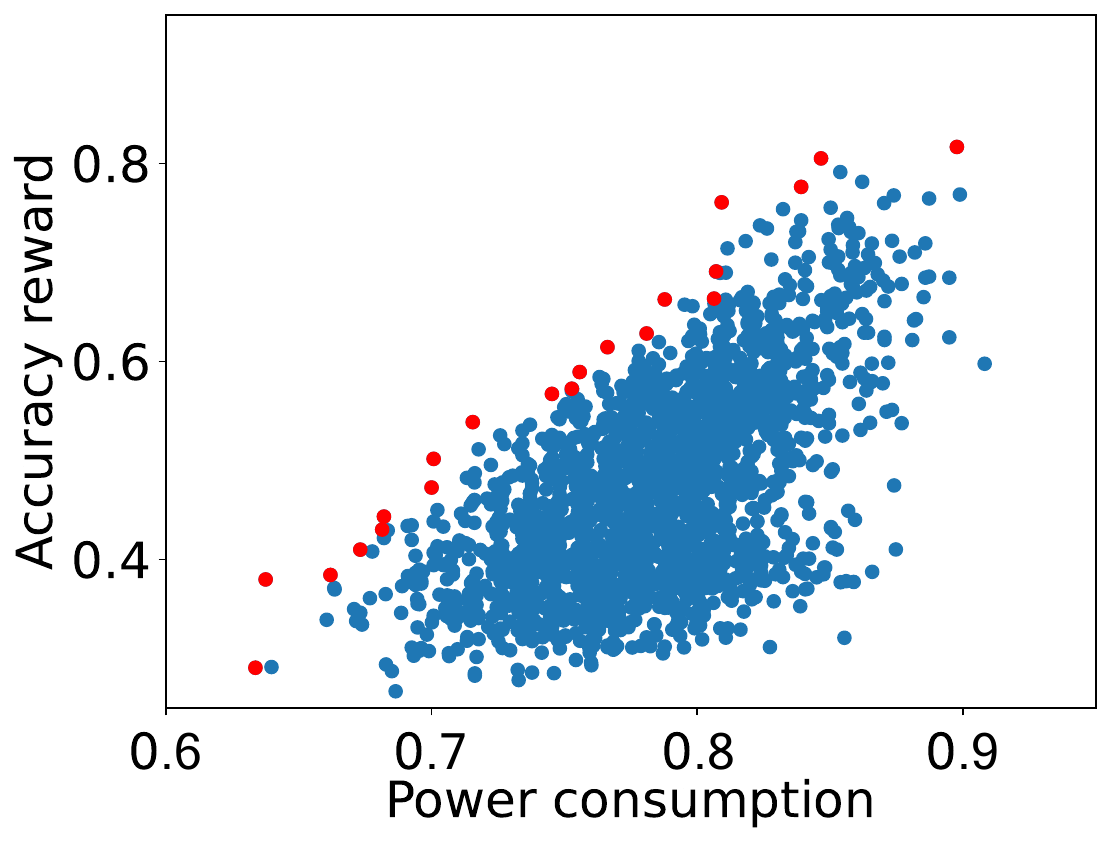}
        \caption{Swin-S}
    \end{subfigure}
    \hfill
    \begin{subfigure}{0.24\textwidth}
        \centering
        \includegraphics[width=\linewidth]{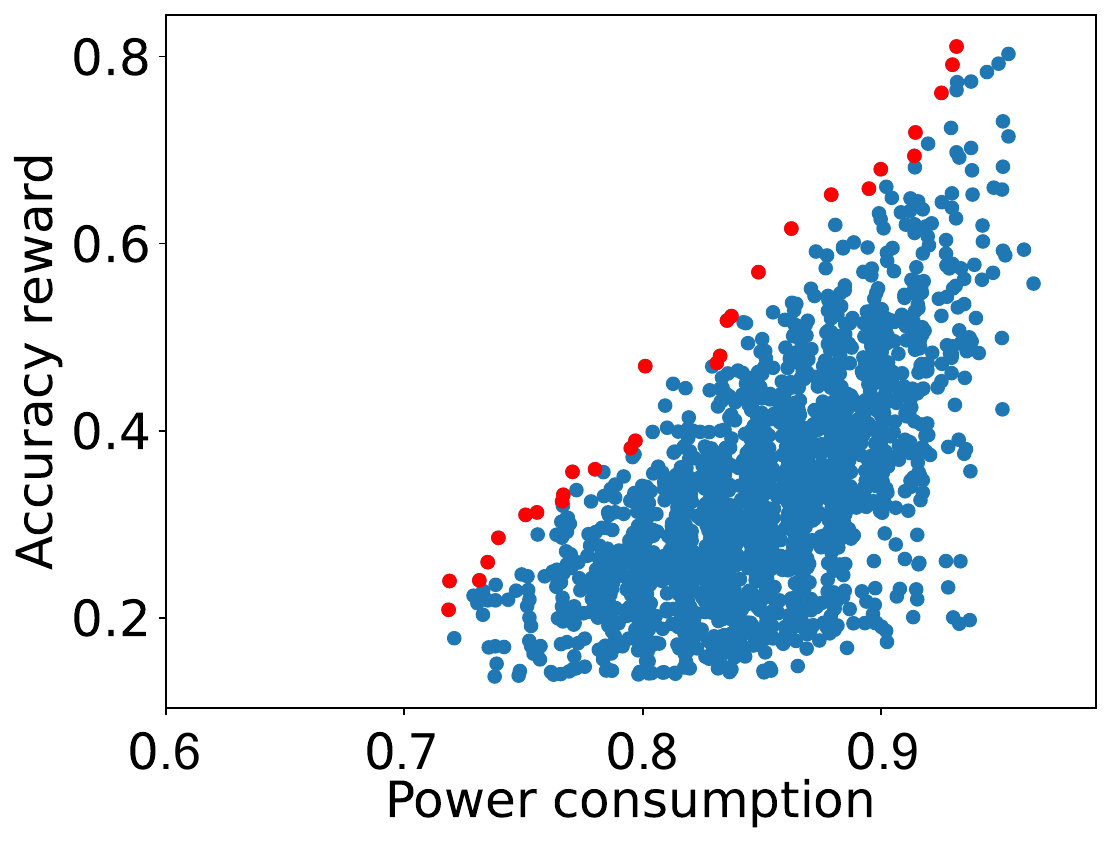}
        \caption{GCViT-XXT}
    \end{subfigure}
    \caption{Scatter plots of hw-driven MCTS using 2000 simulations for $\lambda=1.5$ (top) and $\lambda=0.5$ (bottom) parameter.}
    \label{fig:mcts}
\end{figure*}

\label{subsec:mcts_sims}
\textbf{MCTS simulations}: Finally, we evaluate the use of our hardware-driven MCTS algorithms towards finding the Pareto-optimal curve for accuracy and power. In \figurename~\ref{fig:mcts}, we illustrate the scatter plots from MCTS for every target ViT model using 2000 simulations (power consumption is normalized). The number of simulations shown in the plots is for demonstration and illustrative purposes. In the actual search algorithm, we used approximately 8000 simulations. For a state in the tree to be evaluated accurately, it must be visited a sufficient number of times (to gain the confidence about the statistics). Thus, the number of simulations clearly affects the quality of solutions, but we saw that around 8000 simulations were enough for most models to derive successful results in a reasonable time, as also proven from the convergence of the search algorithm in \figurename~\ref{fig:mcts_convergence}. On average, the exploration time on most models for this setup was about $2$ hours, which is considered very acceptable, especially when compared with previous works \cite{Yao2021, alwann}. The user has the flexibility to tune it according to their requirements; however, further exploration yielded minimal results on the accuracy-power trade-off. Each acquired Pareto (in red) represents the knowledge learned by the system towards finding the optimal multiplier configuration and it's basically the output of the search algorithm. To provide further comparisons, we experiment for two distinct $\lambda$ parameters for the power bias, as described in Algorithm \ref{alg:algorithm} and Equation \ref{eq:policy}. When $\lambda$ is low the model is not penalized for high power consumption, thus there are slightly more solutions/data points with high power consumption in the corresponding scatter plots. We deliberately kept the difference between the two $\lambda$ values relatively narrow as significant changes could drastically alter the reward landscape, and thus disrupt the MCTS search algorithm. For example, very high or very low $\lambda$ values could result in suboptimal models with poor performance as good solutions might be overlooked by the search algorithm. Last, as we evaluated our approach on many models, we chose a $\lambda$ value that could capture the specific dynamics of every model as much as possible. Experimenting with different $\lambda$ parameters or exploration-exploitation ratios could potentially yield marginally improved results; however, our experiments serve as a strong indication of the successful application of the proposed search algorithm and parameters.

\begin{figure}
  \centering
    \caption*{\footnotesize \hspace{-0.01cm}  \ \ \ (a) ViT-S  \hspace{3.3cm}
     (b) DeiT-S  }
    \vspace{-0.3\baselineskip}   
  \includegraphics[width=0.24\textwidth]{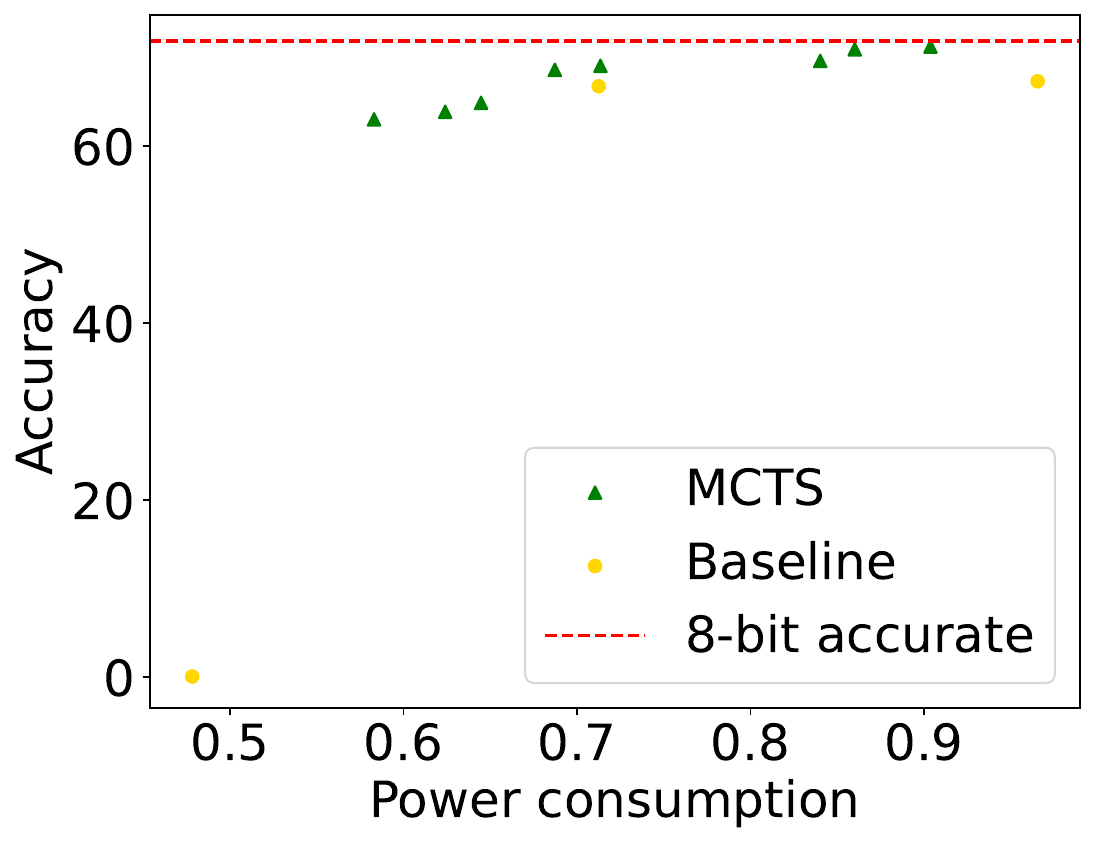}
  \hfill
  \includegraphics[width=0.24\textwidth]{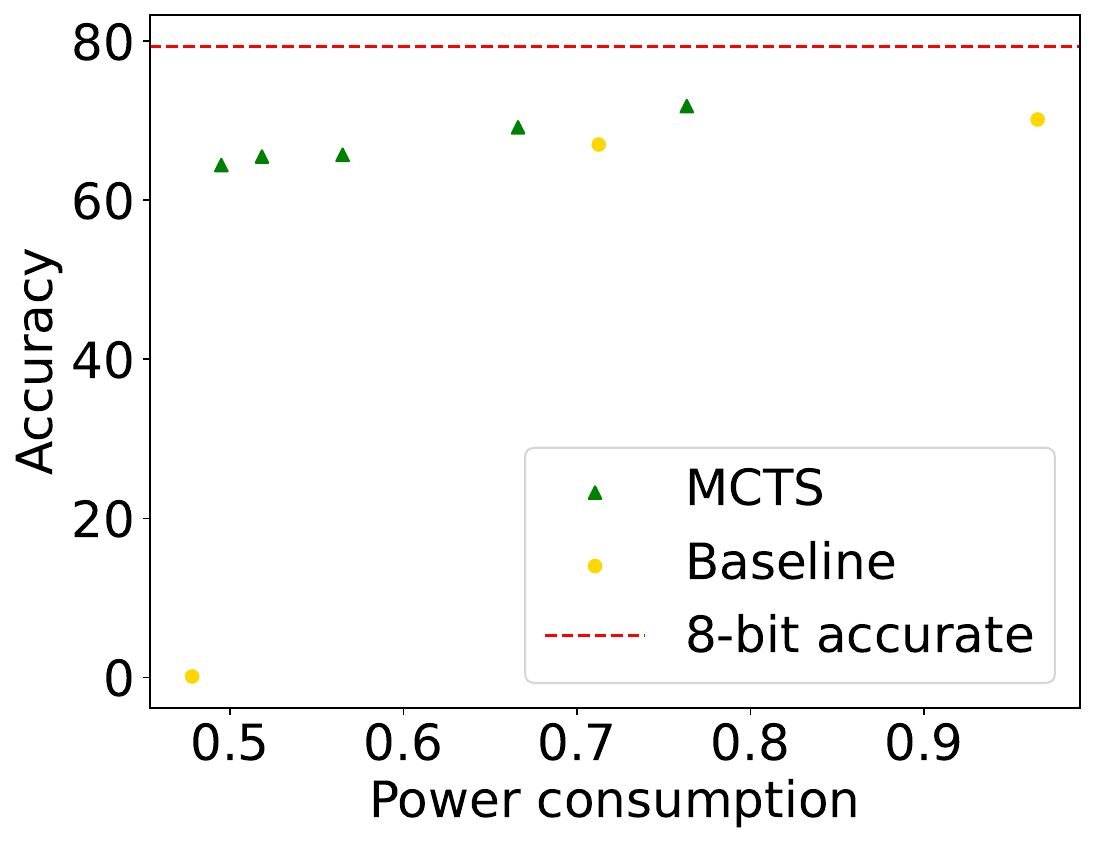}
    \caption*{\footnotesize \hspace{0.4cm}  \ \ \ (c) Swin-S  \hspace{2.6cm}
    \ (d) GCViT-XXT }
     \vspace{-0.3\baselineskip} 
  \includegraphics[width=0.24\textwidth]{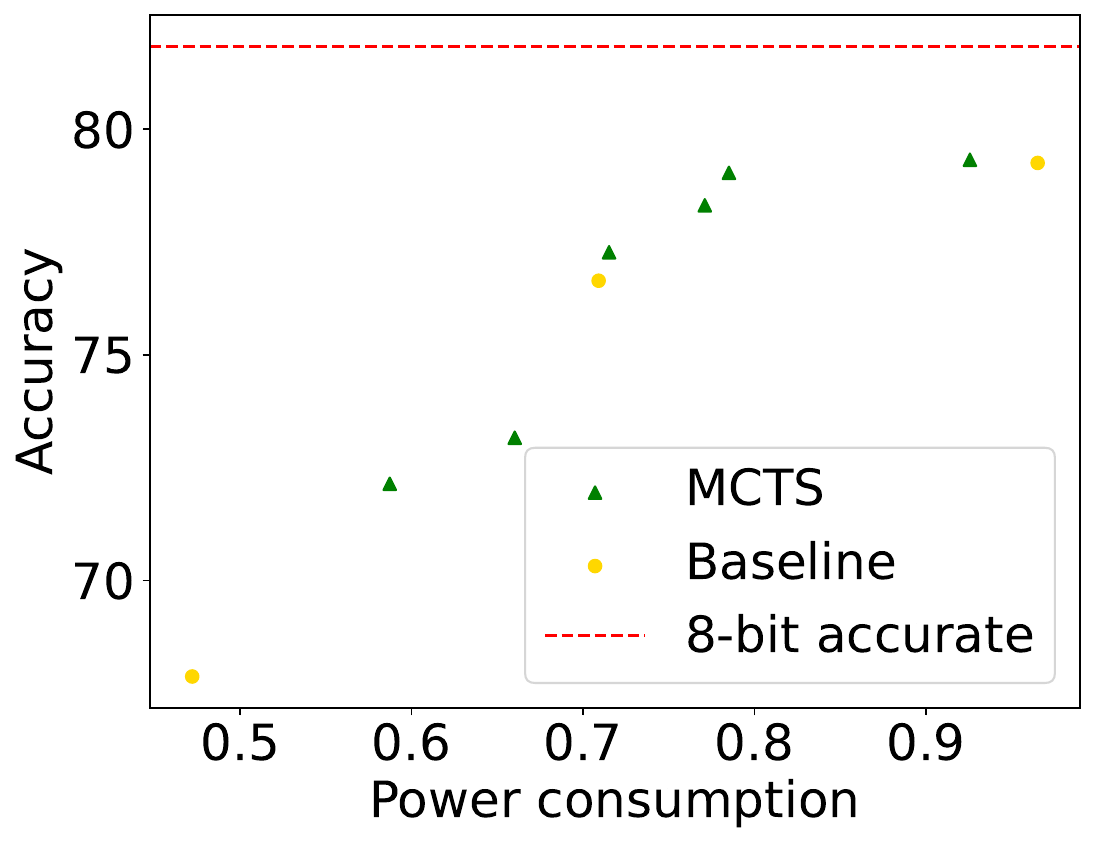}
  \hfill
  \includegraphics[width=0.24\textwidth]{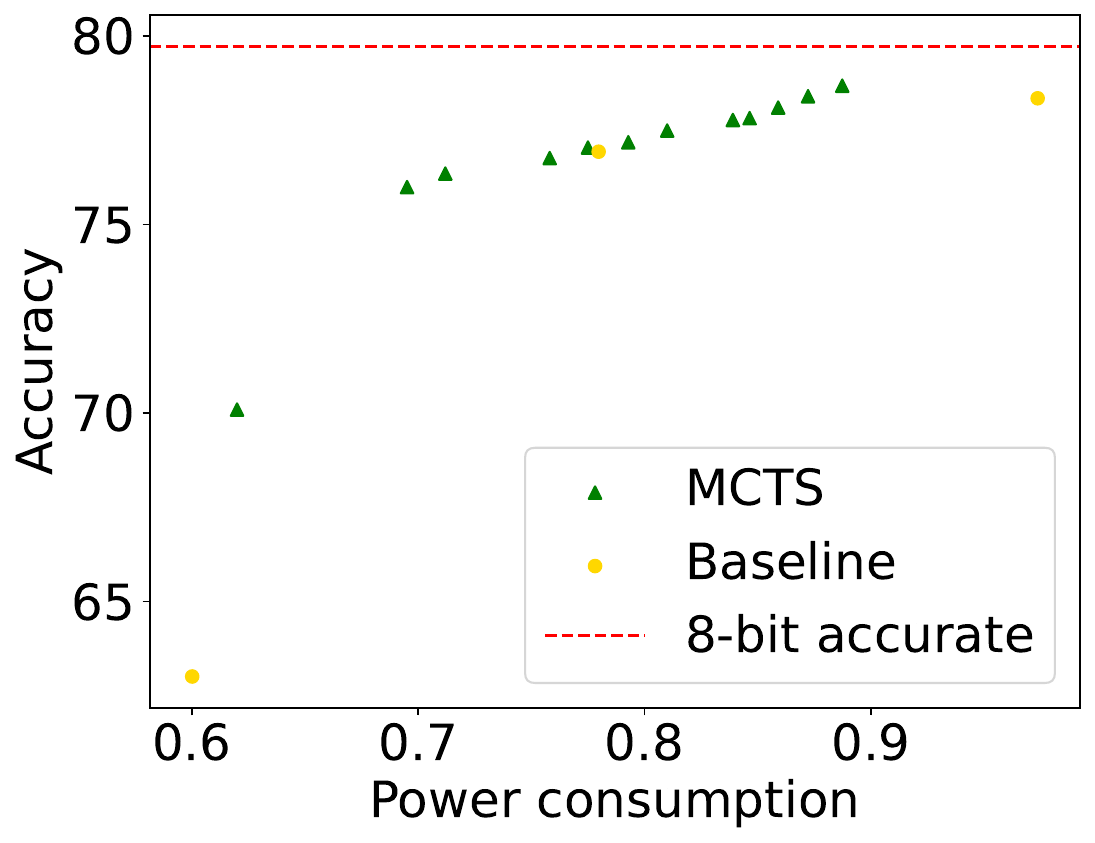}
  
  \vspace{0.5\baselineskip}   
  \caption{MCTS-optimal solutions based on real accuracy (green) and baseline approximate solutions (yellow).}  
  \label{fig:mcts_acc_results}
  \vspace{-2ex}  
\end{figure}

In \figurename~\ref{fig:mcts_acc_results}, we summarize the solutions found with the baseline approximation of the models (in yellow), obtained from Table \ref{tab:acc_results}, along with the proposed optimal solutions obtained by our MCTS-based approach (in green). As baseline solutions we define the accuracy/power data points from Table \ref{tab:acc_results} in which approximation is uniformly applied to all layers of the model without much customization.
The corresponding scatter plots of \figurename~\ref{fig:mcts_acc_results} illustrate the practical observation of the accuracy/power trade-off, with normalized power consumption along the x-axis and actual measured accuracy along the y axis. Notably, the study highlights the superior performance of our hardware-driven Monte Carlo tree search algorithm, particularly in achieving a more advantageous balance between power consumption and accuracy, as well as giving an increased flexibility from the designer's perspective in choosing a wider range of accuracy-power approximate solutions. 

The green data points of \figurename~\ref{fig:mcts_acc_results} are derived from the large exploration of the parameter space using MCTS, depicted in \figurename~\ref{fig:mcts}, focusing only on the optimal solutions achieved through this algorithm (red points in \figurename~\ref{fig:mcts}). These data points are evaluated on the whole ImageNet validation dataset and the Pareto points are depicted in the corresponding scatter plots of \figurename~\ref{fig:mcts_acc_results} as green triangular markers.Re-training was required  for these solutions to fine-tune the final accuracy. Typically, epochs ranged from 5 to 15, with many instances employing an early stopping method to halt re-training once the loss converged.
 The baseline solutions from Table \ref{tab:acc_results}, for each approximate multiplier, are deliberately included to provide a benchmark for evaluating the effectiveness of the proposed solutions using the MCTS search algorithm. 
  
The scatter plots of \figurename~\ref{fig:mcts_acc_results} visually highlight the trade-offs between power consumption and accuracy. Our solutions (green points) mainly form the Pareto-front in each subplot (i.e., deliver the best best possible compromise between the two objectives, accuracy and power.)
    For similar accuracy as the baseline approximate solutions (within $\sim$ \!\!\!\!\!\!\ $1\%$ maximum difference), MCTS delivers on average solutions with $\sim \!\!\!\!\ 21\%$ lower power. Additionally, our approach yields a considerable number of Pareto solutions in all models, providing the designer with a fine-grained set of choices. These choices aim to deliver an effective balance between the two objectives, power and accuracy, while minimizing exploration time. In contrast, the baseline approximation offers a more limited selection.
   

In general, our MCTS search facilitates obtaining results from the Pareto-curve, tailored to meet the designer's requirements, ultimately providing a more nuanced trade-off between accuracy and power compared to the baseline approximation. Our specific application requirements included constraints on power consumption, accuracy thresholds, and exploration time for the MCTS algorithm. However, additional criteria can be incorporated depending on the system under consideration. To the best of our knowledge, our work is pioneering in analyzing the impact of approximate multipliers on ViT models and offering a framework that: a) evaluates inference accuracy with reasonable speed, b) supports approximation-aware ViT re-training, and c) achieves a detailed accuracy-power trade-off in the DSE of ViT to approximation mapping.

Despite the considerable savings reported in \figurename~\ref{fig:mcts_acc_results}, higher power savings for similar accuracy loss have often been reported in approximate CNNs \cite{AxDNNsurvey}.
Hence, additional research is required for ViT models and/or dedicated approximate multipliers/approximation techniques are needed.
Our work lays the groundwork for efficiently exploring the balance of power/accuracy in ViTs, enabling designers to quickly identify optimal solutions, facilitating effective design space exploration and co-design approaches. To the best of our knowledge, TransAxx is the only tool available that can assist the design and evaluation of approximate multipliers in ViTs.

\section{Limitations and Perspective}\label{sec:perspective}
\begin{enumerate}[left=0pt]
\item 
TransAxx is not designed to create approximate multipliers; rather, its purpose is to evaluate their impact on ViT accuracy and facilitate layer-to-approximation mapping using MCTS.
Targeting the design of approximate multipliers, particularly for DNNs, some works focus on the error metrics (e.g., variance or root mean square error) that should be optimized when designing approximate multipliers for DNN accelerators~\cite{weightoriented, 8714977, 8863138} and others on proposing error correction mechanisms~\cite{weightoriented,tfapprox,axconvar, 8727537}. 
Nevertheless, none of these works supports Transformers and ViT models in particular.
TransAxx, however, can be leveraged to derive similar statistical insights for approximate Transformers.
\textit{Ultimately, TransAxx complements hardware EDA tools by addressing the gap in accuracy evaluation for ViT models under approximate multiplications, providing designers with the only solution for efficiently exploring the respective accuracy-hardware savings design space.}

\item TransAxx is the only tool supporting ViT model accuracy under approximate multiplications.
Though, establishing acceptable quality (approximation) levels for a desired application is entirely application-dependent.
The correlation between Transformer accuracy and application quality may not always be straightforward and should be explored by the user for each specific use case. 
Varying applications may require optimization of different error metrics. 
Moreover, the challenge of defining acceptable levels of approximation lies in the fact that error tolerance varies widely with application as well as, in many cases, external environmental parameters.
While TransAxx is designed to emulate inference with approximate multipliers, it operates in Python and can therefore be integrated with any other application or framework to evaluate the impact of approximate multiplications not only on the accuracy of Transformers but also on the quality of the overall application.
Hence, \textit{TransAxx can seamlessly complement existing frameworks and approximate computing tools to define acceptable levels of quality according to the application by providing fast and effortless approximate ViT emulation}.

\item As discussed in Section~\ref{subsec:axmults}, TransAxx does not natively support VOS-approximated multipliers.
However, VOS-based circuits are widely used due to their quadratic power savings~\cite{vossim,Zervakis:TCASII2019,Senobari:Access2024,Bahoo:TETC2023,Zervakis:ISLPED2015}, spanning DSP and ML applications~\cite{Zervakis:TCASII2019}.
Particularly, voltage scaling in approximate DNN circuits remains an active research area~\cite{Senobari:Access2024,XNVDLA,GreenTPU,axthermal}
In~\cite{XNVDLA}, authors store VOS-approximated multiplier outputs in LUTs for all input combinations.
If these LUTs rely only on the current multiplicand values, they can be seamlessly reused in TransAxx.
\textit{Statistical approaches, such as~\cite{Senobari:Access2024}, appear more practical for integrating VOS support into TransAxx} compared to LUT-based methods, which might require excessive memory and execution time, as VOS-LUTs could become three- or four-dimensional if TransAxx must account for both current and previous inputs.
This statistical integration aligns well with TransAxx’s function-based mode, though users must carefully consider factors like DNN mapping, dataflow, accelerator microarchitecture, and PE arrangement.
Unlike logic and algorithmic approximations, where TransAxx directly replaces accurate products with approximate ones, VOS errors depend on input sequences that might be affected by such parameters.
For instance,~\cite{Senobari:Access2024} incorporates the number of PEs in a column into its error model.
Nevertheless, in TransAxx, the approximate multiplication function can be seamlessly extended, allowing users to define the desired functionality.
It is important to note, however, that such parameters (e.g., DNN mapping, dataflow, number of PEs) are not required when emulating logic and algorithmic approximations in TransAxx.
Moreover, statistical-based approaches will offer a good enough estimation of the VOS error with some confidence, whereas for logic/algorithmic approximations, TransAxx computes the error precisely.

\item In the LUT-based approach, TransAxx precomputes and stores all possible approximate products. However, these LUTs are read-only and must be generated externally before execution. While new LUTs cannot be modified within TransAxx, the framework supports just-in-time compilation, enabling smooth integration of updated LUTs.
This read-only nature does not restrict flexibility since (i) most approximate multipliers are supported (see Section~\ref{subsec:axmults}), (ii) users typically select multipliers from existing libraries, e.g.,~\cite{alwann}, and (iii) in co-design scenarios, new multipliers can be integrated by regenerating their LUTs.
Since multiplier modifications and hardware evaluations occur outside TransAxx, generating new LUTs incurs minimal overhead and can overlap with hardware assessment.
Moreover, the approximation-aware ViT training feature of TransAxx will also help to potentially recover any accuracy loss. 
\end{enumerate}

\section{Conclusion}

In this paper we introduced TransAxx, an end-to-end framework built on top of PyTorch library, that is designed for seamless and fast evaluation and re-training of approximate Vision Transformers. TransAxx is accelerated by leveraging GPU hardware without significant overhead compared with the native execution. Additionally, we introduce a novel methodology for searching the design space of approximate ViT models using a hardware-driven MCTS-based algorithm. Our findings demonstrate the capability to achieve substantial gains in both accuracy and power, according to the preferences of the designer, all within a short timeframe. Towards contributing fundamentally in the software-hardware ecosystem, TransAxx will be made available open-source.

\bibliographystyle{IEEEtran}
\bibliography{references}

 \vspace{-1cm}

\begin{IEEEbiography}
[{\includegraphics[width=1in,height=1.25in,clip,keepaspectratio]{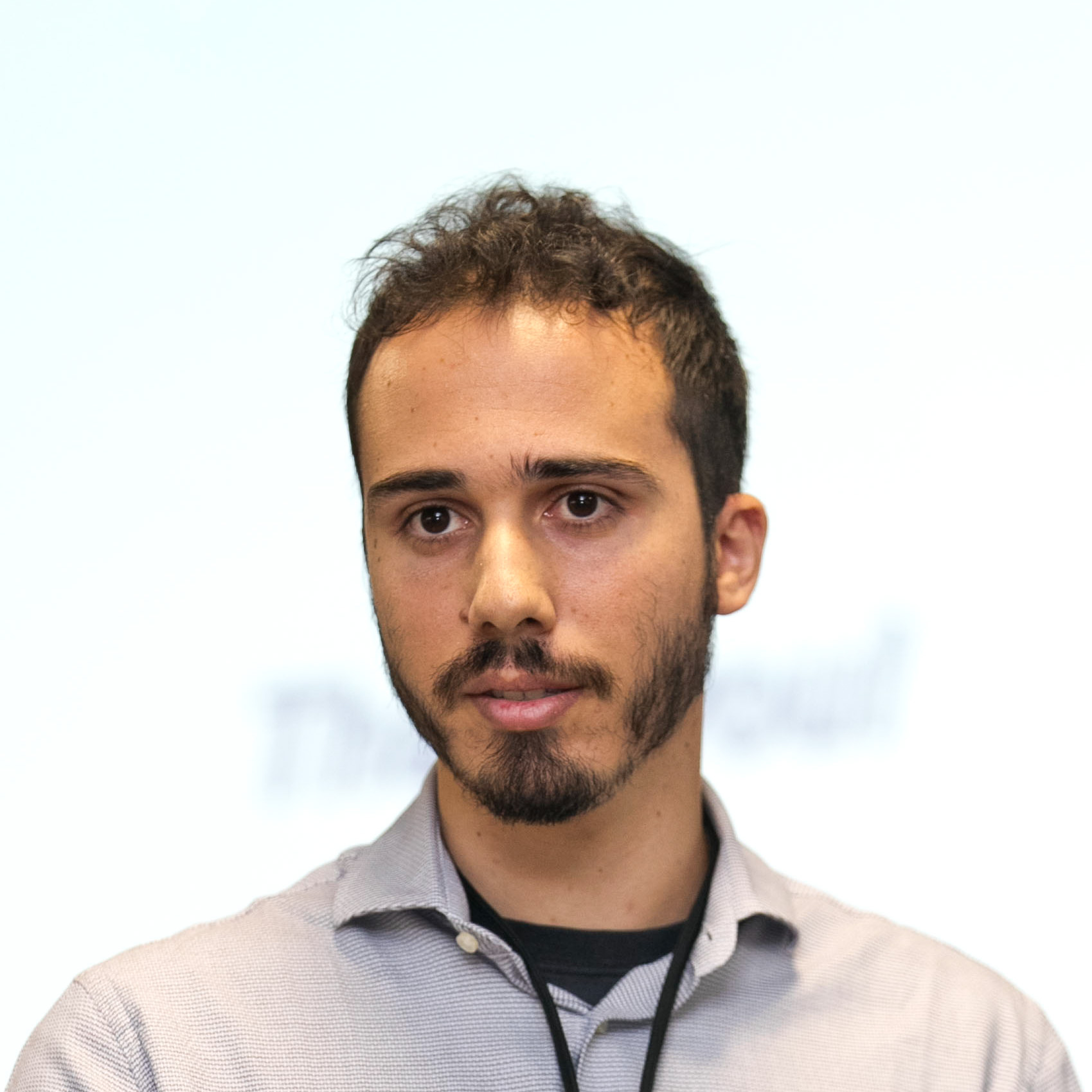}}]{Dimitrios Danopoulos} received his diploma in Electrical and Computer Engineering from the National Technical University of Athens (NTUA) in 2018. He is currently a PhD canditate at the School of Electrical and Computer Engineering of the National Technical University of Athens. His research area involves the development and hardware acceleration of Machine Learning and Deep Learning applications. He has also contributed to various European projects by participating in the research teams.
\end{IEEEbiography}

\vspace{-1cm}

\begin{IEEEbiography}[{\includegraphics[width=1in,height=1.25in,clip,keepaspectratio]{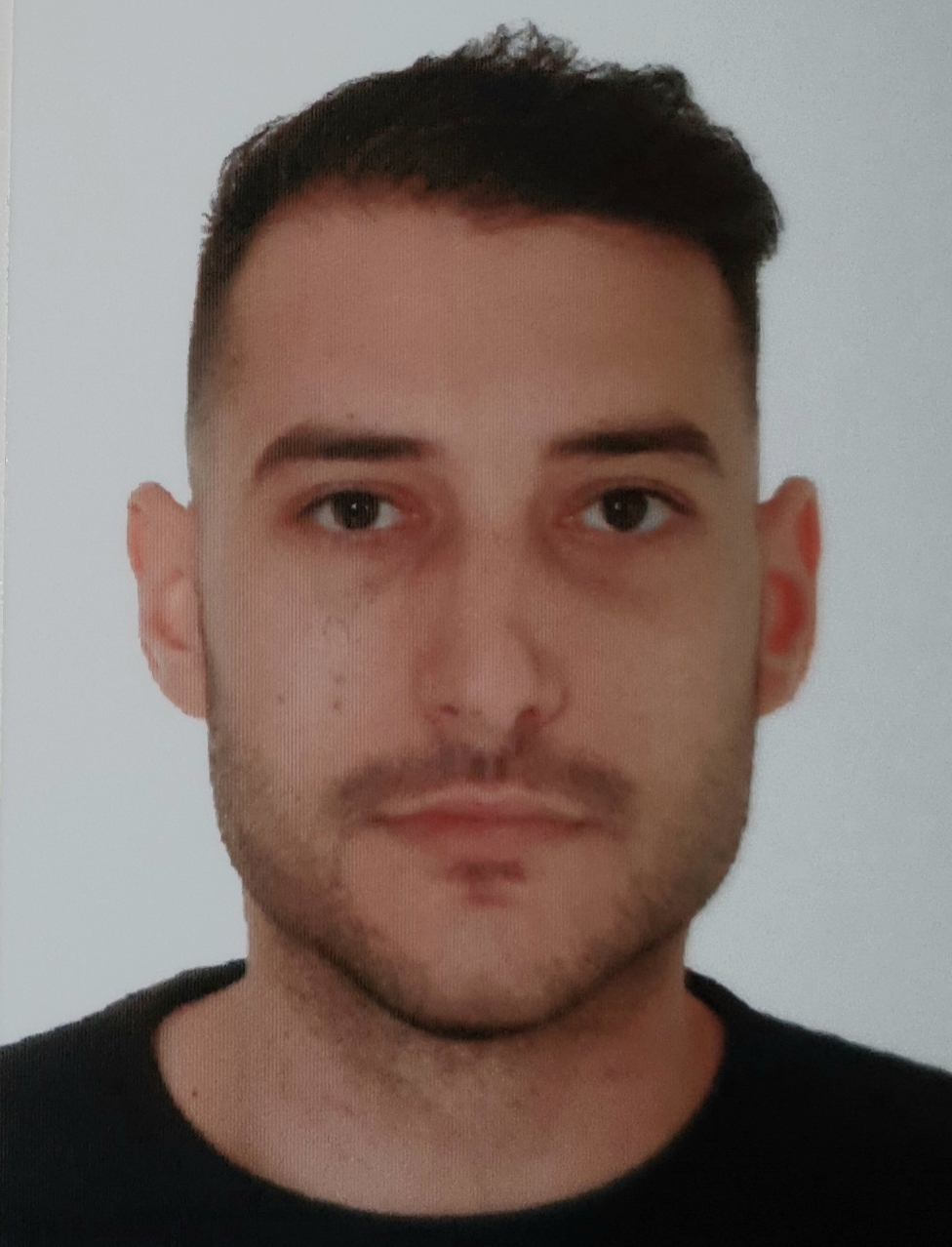}}] {Georgios Zervakis}  is an Assistant Professor at the University of Patras. Before that he was a Research Group Leader at the Chair for Embedded Systems (CES), at the Karlsruhe Institute of Technology (KIT) from 2019 to 2022. He received the Diploma and Ph.D. degrees from the School of Electrical and Computer Engineering (ECE), National Technical University of Athens (NTUA), Greece, in 2012 and 2018, respectively. Dr. Zervakis serves as a reviewer in many IEEE and ACM journals and is also a member of the technical program committee of several major design conferences. He has received one best paper nomination at DATE 2022. His main research interests include low-power design, accelerator microarchitectures, approximate computing, and machine learning.
\end{IEEEbiography}

\vspace{-1cm}

\begin{IEEEbiography}
[{\includegraphics[width=1in,height=1.25in,clip,keepaspectratio]{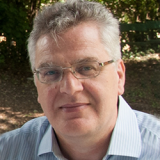}}]{Dimitrios Soudris} received his diploma and PhD in Electrical Engineering from the University of Patras in 1987 and 1992, respectively. He worked as a Lecturer, Assistant and Associate Professor in the Department of Electrical and Computer Engineering, Democritus University of Thrace for thirteen years from 1995 until 2008. Currently, he works as a Full Professor in the School of Electrical and Computer Engineering, National Technical University of Athens. He has published more than 550 papers in international journals and conferences, holds two international patents and his research has been cited \textgreater 6000 times. Also, he is the author and editor of ten books Kluwer and Springer. He is Director of Microprocessor and Digital systems Lab (MicroLab). Last, he has been project coordinator and/or and principal investigator in numerous research \& development projects (\textgreater 60) funded by the European Commission, ENIAC-JU, European Space Agency, the Greek Government and the European and Greek Industry. 
\end{IEEEbiography}

\begin{IEEEbiography}[{\includegraphics[width=1in,height=1.25in,clip,keepaspectratio]{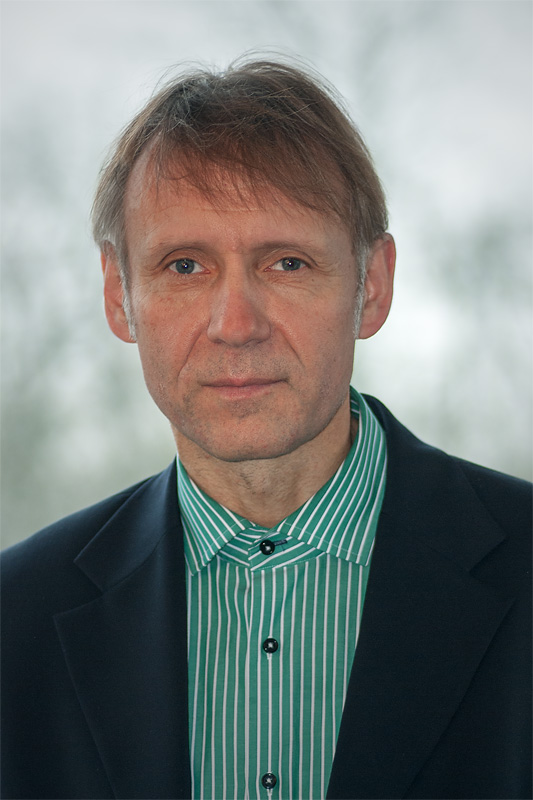}}]{J\"org Henkel} (M'95-SM'01-F'15) is the Chair
Professor for Embedded Systems at Karlsruhe
Institute of Technology. Before that he was a research staff member at NEC Laboratories in
Princeton, NJ.
He has received six best paper awards from, among others, ICCAD, ESWeek and DATE.
For two terms he served as the Editor-in-Chief for the ACM Transactions on Embedded Computing Systems.
He is currently the Editor-in-Chief of the IEEE Design\&Test Magazine and is/has been Associate Editor for major ACM and IEEE Journals.
He has led several conferences as a General Chair incl. ICCAD, ESWeek and serves as Steering Committee chair/member for leading conferences and journals for embedded and cyber-physical systems. Prof. Henkel coordinates the DFG program SPP 1500 ``Dependable Embedded Systems'' and is a site coordinator of the DFG TR89 collaborative research center ``Invasive Computing''. He is the chairman of the IEEE Computer Society, Germany Chapter, and a Fellow of the IEEE.
\end{IEEEbiography}

\end{document}